\documentclass[runningheads]{llncs}

\usepackage{eccv}

\usepackage{eccvabbrv}

\usepackage{graphicx}
\usepackage{booktabs}
\usepackage[numbers,sort&compress]{natbib}
\usepackage{pifont}
\usepackage{multirow}

\usepackage{amsmath}
\usepackage{amssymb}
\usepackage{mathtools}

\usepackage{subcaption}
\definecolor{darkred}{rgb}{0.6148, 0., 0.}
\usepackage{array}

\newcolumntype{?}{!{\vrule width 1pt}}
\newcolumntype{|}{!{\vrule width .5pt}}
\usepackage[accsupp]{axessibility}  

\usepackage[table]{xcolor}
\usepackage{xspace}
\newcommand{\our}{\texttt{TSEmbed}\xspace}
\usepackage{tikz}
\usepackage{pgfplots}
\definecolor{graybg}{HTML}{F2F2F2} 

\usepackage[pagebackref,breaklinks,colorlinks,citecolor=eccvblue]{hyperref}

\usepackage{orcidlink}

\begin{document}


\title{TSEmbed: Unlocking Task Scaling in Universal Multimodal Embeddings}

\author{Yebo Wu\inst{1} \and
Feng Liu\inst{2} \and
Ziwei Xie\inst{2}  \and 
Zhiyuan Liu\inst{2} \and 
Changwang Zhang\inst{2} \and \\ 
Jun Wang\inst{2} \and 
Li Li\inst{1}}

\authorrunning{Y. Wu et al.}


\institute{State Key Laboratory of IOTSC, University of Macau \and
OPPO Research Institute \\
\email{\{yc37926,llili\}@um.edu.mo}, 
\email{\{liufeng4hit,junwang.lu\}@gmail.com}\\
\email{\{xieziwei,liuzhiyuan1\}@oppo.com},
\email{changwangzhang@foxmail.com}}
\maketitle

\begin{abstract}
    Despite the exceptional reasoning capabilities of Multimodal Large Language Models (MLLMs), their adaptation into universal embedding models is significantly impeded by task conflict. To address this, we propose \our, a universal multimodal embedding framework that synergizes Mixture-of-Experts (MoE) with Low-Rank Adaptation (LoRA) to explicitly disentangle conflicting task objectives. Moreover, we introduce Expert-Aware Negative Sampling (EANS), a novel strategy that leverages expert routing distributions as an intrinsic proxy for semantic similarity. By dynamically prioritizing informative hard negatives that share expert activation patterns with the query, EANS effectively sharpens the model's discriminative power and refines embedding boundaries. To ensure training stability, we further devise a two-stage learning paradigm that solidifies expert specialization before optimizing representations via EANS. \our achieves state-of-the-art performance on both the Massive Multimodal Embedding Benchmark (MMEB) and real-world industrial production datasets, laying a foundation for task-level scaling in universal multimodal embeddings.

    
\end{abstract}

\section{Introduction}
\label{sec:introduction}

Multimodal embedding models~\citep{jiang2024e5,zhang2025notellm} project heterogeneous inputs into a shared semantic manifold, underpinning a wide range of downstream applications 
such as image-text retrieval~\citep{cao2022image,song2025comprehensive,xu2024copyrightmeter}, retrieval-augmented generation (RAG)~\citep{zhao2026retrieval,yu2024evaluation}, and multimodal recommendation~\citep{ye2025harnessing,liu2024multimodal}.
While foundational vision-language models such as CLIP~\citep{radford2021learning} and SigLIP~\citep{tschannen2025siglip} have established robust cross-modal alignment, their dual-encoder architectures inherently process visual and textual streams in isolation. This late-fusion paradigm creates a structural barrier to deep cross-modal interaction, severely constraining their performance on complex tasks~\citep{xu2025bridging,xu2025videoeraser} that require fine-grained, inter-modal reasoning.


The rapid evolution of Multimodal Large Language Models (MLLMs), exemplified by GPT-4V~\citep{yang2023dawn} and Qwen-VL~\citep{wang2024qwen2}, has catalyzed a promising paradigm shift. 
Pioneering works such as VLM2VEC~\citep{jiang2024vlm2vec} repurpose these generative architectures as universal embedding models, harnessing their expansive world knowledge and compositional reasoning for multimodal representation learning. 
To further enhance their representational capacity, subsequent research has coalesced around three  directions: (1) Data Synthesis~\citep{zhang2024magiclens,li2026qwen3} to overcome the scarcity of high-quality supervised data; (2) Hard Negative Mining~\citep{lan2025llave,xue2025improve} to refine contrastive boundaries; and (3) Reasoning Elicitation~\citep{cui2025think,jiang2026embed} to extract deeper semantic logic via Chain-of-Thought or reinforcement learning.
However, these strategies overlook a fundamental bottleneck: \textbf{task conflict}. Forcing diverse semantic objectives into a monolithic parameter space inevitably induces severe gradient interference, significantly degrading model performance.

To empirically quantify the impact of task conflict, we compare the performance of individual task-specific training against unified joint training (VLM2VEC) on the MMEB.
As illustrated in Figure~\ref{fig:motivation}, the jointly trained model consistently underperforms its task-specific counterparts across all four meta-task categories. Notably, on VQA, the VLM2VEC suffers severe performance drops of 15.1\% and 13.1\% for the 2B and 7B models, respectively. This performance gap confirms that a shared parameter space struggles to satisfy competing optimization objectives, fundamentally constraining the representational capability of MLLMs.

\definecolor{color_task}{HTML}{FF9999} 
\definecolor{color_vlm}{HTML}{99CCFF}  
\definecolor{text_gray}{HTML}{444444}

\begin{figure}[t]
\centering

\begin{minipage}{0.49\textwidth}
\centering
\begin{tikzpicture}
  \begin{axis}[
    font=\scriptsize,
    ymajorgrids,
    grid style={dashed, gray!30},
    xtick pos=left,
    ytick pos=left,
    legend style={at={(0.3,0.65)}, anchor=south, legend columns=1, font=\tiny, draw=none, fill=none},
    ybar=1pt, 
    enlarge x limits=0.15,
    height=0.67\linewidth, 
    width=\linewidth, 
    bar width=1.2em, 
    ylabel={hit@1 (\%)},
    ylabel style={yshift=-1em, font=\small\boldmath},
    symbolic x coords={1, 2, 3, 4, 5, 6},
    xtick=data,
    ymin=40, ymax=100,
    ytick={40, 50, 60, 70, 80, 90, 100},
    xticklabels={Classification, VQA, Retrieval, Grounding},
    xticklabel style={yshift=-0.2em, font=\tiny, rotate=0, anchor=north}, 
    nodes near coords,
    every node near coord/.append style={
        font=\tiny, 
        /pgf/number format/fixed, 
        /pgf/number format/precision=1,
        yshift=0pt
    }
  ]

    \addplot[fill=color_task, draw=black!70, line width=0.6pt] 
    coordinates {(1,68.2) (2,64.5) (3,68.9) (4,86.2)};
    \addlegendentry{Task-Specific}

    \addplot[fill=color_vlm, draw=black!70, line width=0.6pt] 
    coordinates {(1,59.0) (2,49.4) (3,65.4) (4,73.4)};
    \addlegendentry{VLM2VEC}

    \def\diffbar#1#2#3{
        \node[text=red!70!black, font=\tiny\bfseries, yshift=10pt] at (axis cs:#1,#2) {$\downarrow$#3};
    }
    
    \diffbar{1}{68.2}{9.2}
    \diffbar{2}{64.5}{15.1}
    \diffbar{3}{68.9}{3.5}
    \diffbar{4}{86.2}{12.8}

  \end{axis}
\end{tikzpicture}
\centerline{\small (a) Qwen2-VL-2B.} 
\end{minipage}%
\hfill 
\begin{minipage}{0.49\textwidth}
\centering
\begin{tikzpicture}
  \begin{axis}[
    font=\scriptsize,
    ymajorgrids,
    grid style={dashed, gray!30},
    xtick pos=left,
    ytick pos=left,
    legend style={at={(0.3,0.63)}, anchor=south, legend columns=1, font=\tiny, draw=none, fill=none},
    ybar=1pt,
    enlarge x limits=0.15,
    height=0.67\linewidth,
    width=\linewidth,
    bar width=1.2em,
    ylabel={hit@1 (\%)},
    ylabel style={yshift=-1em, font=\small\boldmath},
    symbolic x coords={1, 2, 3, 4, 5, 6},
    xtick=data,
    ymin=50, ymax=104,
    ytick={50, 60, 70, 80, 90, 100},
    xticklabels={Classification, VQA, Retrieval, Grounding},
    xticklabel style={yshift=0em, font=\tiny, rotate=0, anchor=north},
    nodes near coords,
    every node near coord/.append style={
        font=\tiny, 
        /pgf/number format/fixed, 
        /pgf/number format/precision=1,
        yshift=0pt
    }
  ]

    \addplot[fill=color_task, draw=black!70, line width=0.6pt] 
    coordinates {(1,71.5) (2,70.9) (3,75.5) (4,91.7)};
    \addlegendentry{Task-Specific}

    \addplot[fill=color_vlm, draw=black!70, line width=0.6pt] 
    coordinates {(1,62.6) (2,57.8) (3,69.9) (4,81.7)};
    \addlegendentry{VLM2VEC}

    \def\diffbar#1#2#3{
        \node[text=red!70!black, font=\tiny\bfseries, yshift=10pt] at (axis cs:#1,#2) {$\downarrow$#3};
    }
    
    \diffbar{1}{71.5}{8.9}
    \diffbar{2}{70.9}{13.1}
    \diffbar{3}{75.5}{5.6}
    \diffbar{4}{91.7}{10.0}
  \end{axis}
\end{tikzpicture}
\centerline{\small (b) Qwen2-VL-7B.}
\end{minipage}
\caption{\textcolor{darkred}{Impact of task conflict on model performance.} Red annotations ($\downarrow$) indicate the performance drop when switching from task-specific models to the unified VLM2VEC.}
\label{fig:motivation}
\end{figure}
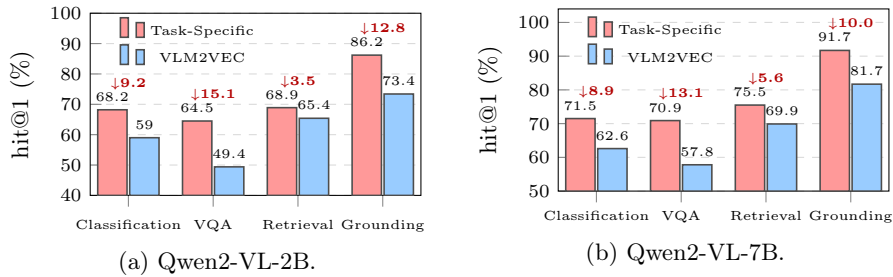

In this paper, we propose \our, a universal multimodal embedding framework that synergizes MoE with LoRA to resolve task conflicts through conditional computation. 
Unlike existing methods that force compromises among divergent objectives, \our partitions the optimization landscape into semantically decoupled subspaces.
By dynamically routing queries to specialized experts, it transforms destructive conflict into collaborative specialization.
Furthermore, we introduce Expert-Aware Negative Sampling (EANS), which leverages the MoE routing distribution as an intrinsic proxy to identify hard negatives, thereby enhancing the model's discriminative power.
To harness the full potential of this synergy, we devise a progressive two-stage learning paradigm that stabilizes expert specialization before deploying EANS to precisely refine the embedding boundaries. Our main contributions are summarized as follows:
\begin{itemize}
    \item We systematically analyze task conflict in universal multimodal embeddings across Spatial, Temporal, and Ecological dimensions, exposing the limitations of monolithic adapters in satisfying divergent semantic objectives.

    \item We propose \our, a novel architecture that synergizes MoE with LoRA to mitigate task conflicts through conditional computation. This design explicitly decouples the optimization landscape, transforming destructive gradient interference into collaborative specialization, thereby establishing a foundation for task-level scaling in universal multimodal embeddings.
    
    \item We introduce Expert-Aware Negative Sampling (EANS), a zero-overhead strategy that leverages intrinsic MoE routing distributions as a proxy for semantic similarity to dynamically up-weight hard negatives. To ensure the reliability of these routing signals, we further devise a progressive two-stage learning paradigm that stabilizes expert specialization prior to EANS refinement, effectively sharpening embedding boundaries.

    \item We conduct extensive evaluations on both the MMEB and real-world production datasets. The results demonstrate that \our consistently achieves state-of-the-art performance across both public benchmarks and practical applications, yielding a notable \textbf{21.87\%} gain in advertising scenarios.

\end{itemize}

\section{Anatomy of Task Conflict in Multimodal Embeddings}
\label{sec:motivation}


To mechanistically understand task conflicts in universal multimodal embeddings, we empirically analyze the MMEB using Qwen2-VL-7B~\citep{wang2024qwen2vlenhancingvisionlanguagemodels}, deconstructing the conflict into three dimensions: Spatial, Temporal, and Ecological.


\begin{figure*}[!t]
    \centering
    \begin{subfigure}{0.32\textwidth}
        \centering
        \includegraphics[width=\linewidth]{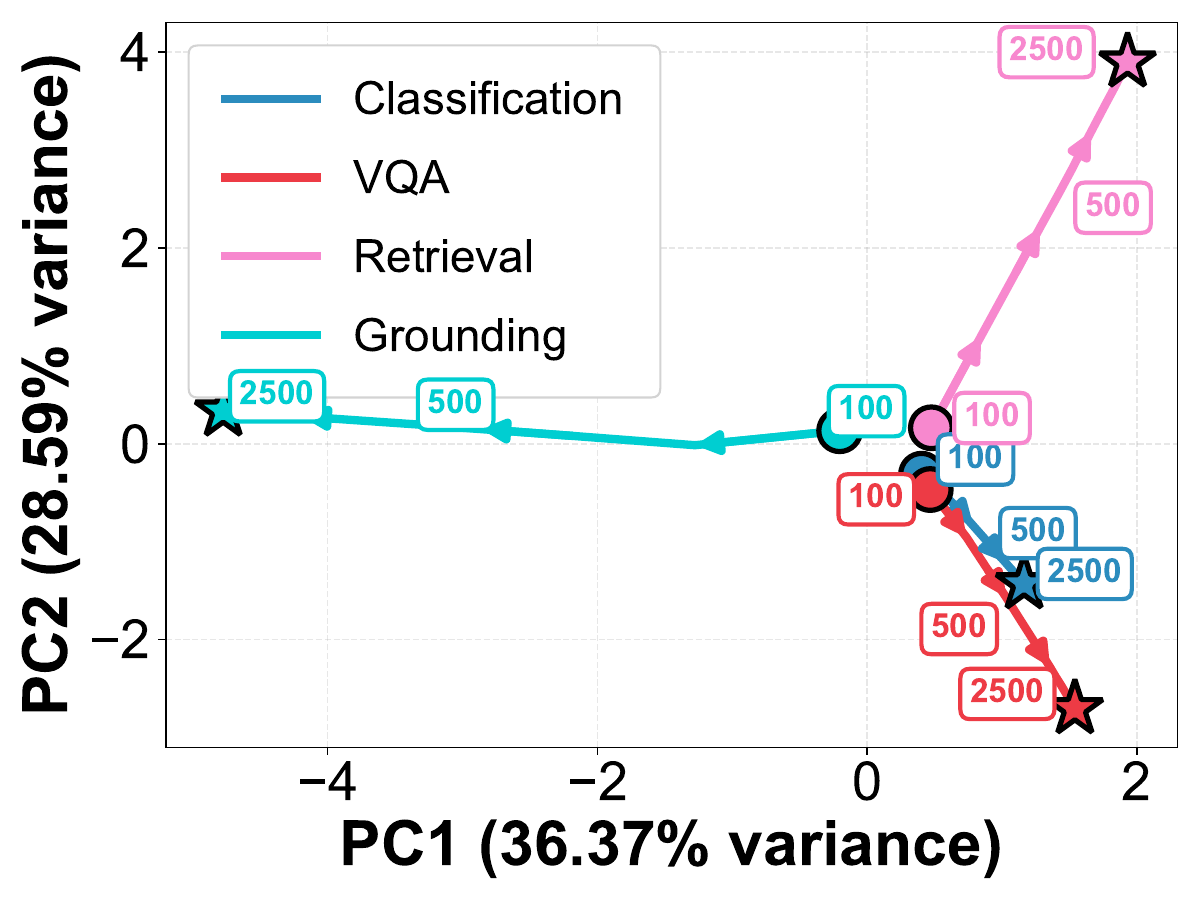}
        \caption{\textbf{Spatial Conflict.}}
    \end{subfigure}
    \begin{subfigure}{0.32\textwidth}
        \centering
        \includegraphics[width=\linewidth]{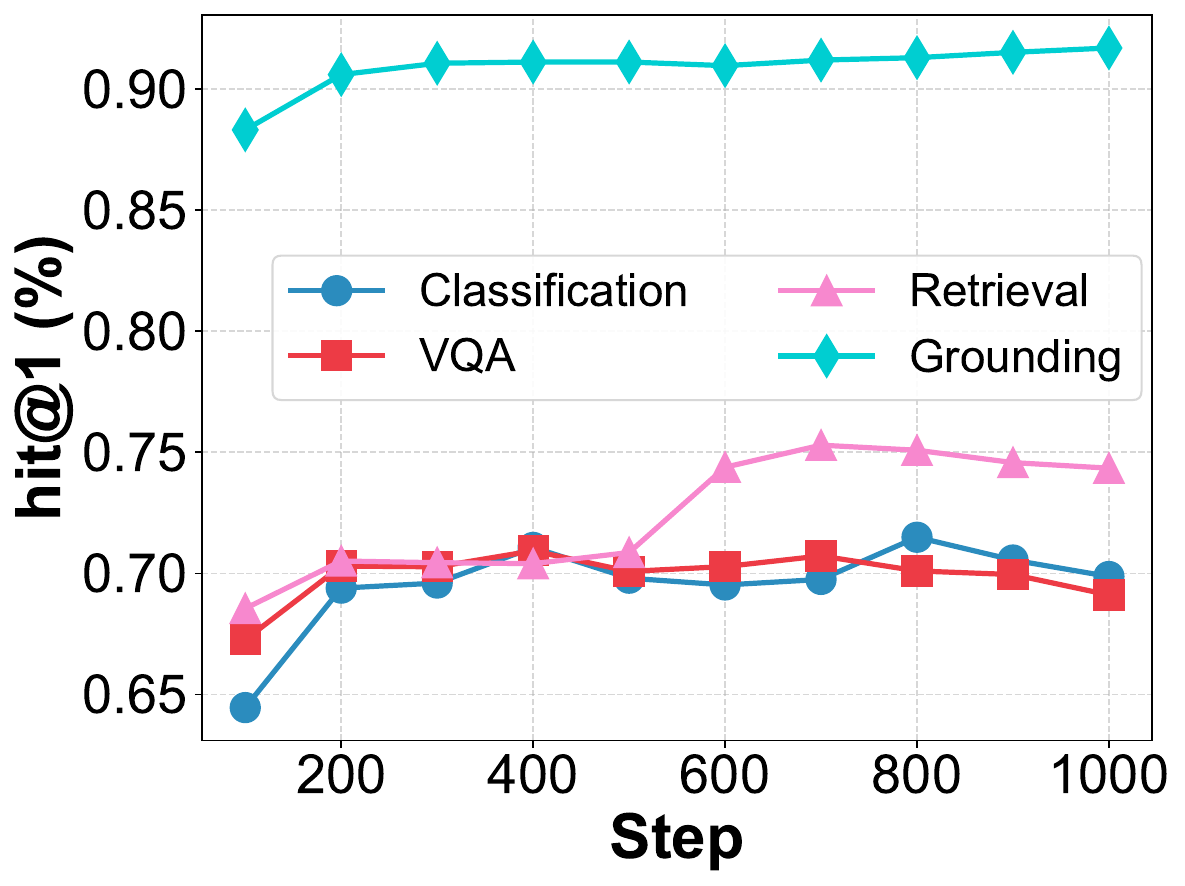}
        \caption{\textbf{Temporal Mismatch.}}
    \end{subfigure}
    \begin{subfigure}{0.32\textwidth}
        \centering
        \includegraphics[width=\linewidth]{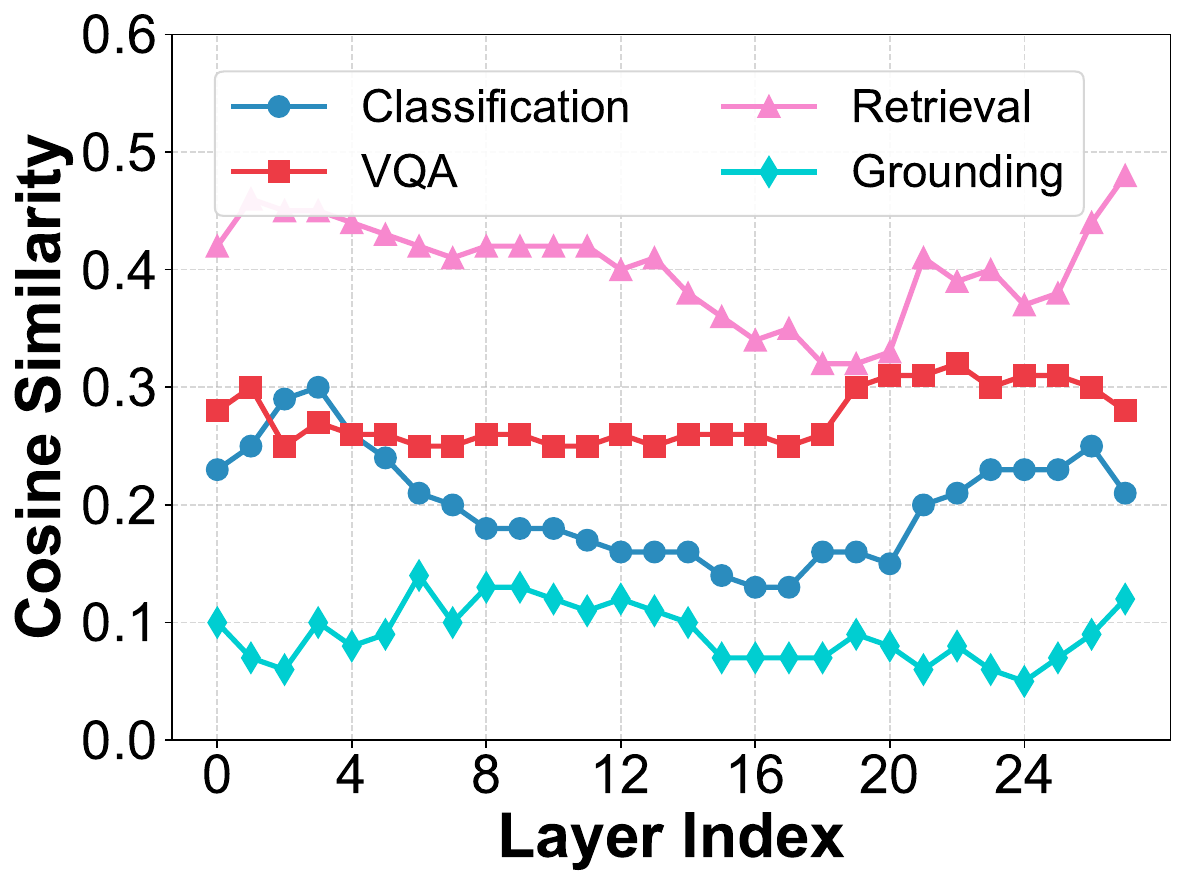}
        \caption{\textbf{Ecological Bias.}}
    \end{subfigure}
    \caption{\textcolor{darkred}{A Multidimensional Anatomy of Task Conflict in Monolithic Adapters.} (a) Divergent optimization trajectories of isolated task-specific adapters. (b) Heterogeneous convergence dynamics of individual tasks during training. (c) Layer-wise cosine similarity between the jointly trained adapter and its isolated counterparts.}
  \label{Fig_motivation1}
\end{figure*}

\subsection{Spatial Dimension: Divergent Gradient Trajectories}
\label{subsec:gradient_divergence}

To uncover the geometric nature of task conflict, we first examine the optimization landscape of isolated tasks. Specifically, we independently train four distinct LoRA adapters, each dedicated to a specific meta-task category. To map the divergence of their optimization paths, we extract model checkpoints at 100-step intervals and project their high-dimensional parameter states into a 2D subspace via PCA~\citep{greenacre2022principal}, which captures a substantial 64.96\% of the total variance.

As illustrated in Figure~\ref{Fig_motivation1}(a), although all task-specific adapters originate from a shared initialization, their training trajectories rapidly splinter into opposing quadrants.
Visual Grounding evolves along the negative PC1 axis, nearly orthogonal to Retrieval's sharp divergence along the positive PC2 axis.
Concurrently, Classification and VQA traverse toward the bottom-right quadrant, in stark geometric opposition to Retrieval.
These highly divergent trajectories confirm that the optimal parameter configurations for distinct tasks reside in entirely disparate regions of the solution space.

\noindent\textit{\textbf{Takeaway I}: The heterogeneous optimization trajectories reveal that the optimal solutions for distinct tasks reside in disjoint parameter spaces.}

\subsection{Temporal Dimension: Heterogeneous Convergence Dynamics}
\label{subsec:convergence_heterogeneity}

Beyond spatial conflicts, we investigate the temporal optimization dynamics of isolated tasks. As illustrated in Figure~\ref{Fig_motivation1}(b), the four meta-tasks exhibit a pronounced temporal misalignment, bifurcating into two divergent patterns. 
Early-converging tasks, such as Visual Grounding and VQA, exhibit rapid initial gains but experience severe degradation as training progresses.
Visual Grounding, for instance, swiftly captures coarse spatial localization, reaching 88.30\% by step 100 and plateauing at 90.58\% by step 200. Conversely, late-converging tasks like Retrieval and Classification demand sustained optimization horizons.


This temporal heterogeneity imposes a strict synchronization bottleneck on monolithic architectures. A single, shared learning schedule cannot simultaneously accommodate these divergent patterns: halting training prematurely to preserve VQA performance leaves Retrieval severely underfitted, while extending the optimization horizon causes early-converging tasks to overfit and degrade.

\vspace{1mm}
\noindent\textit{\textbf{Takeaway II:} Distinct tasks converge at markedly different speeds, making a single shared adapter incapable of synchronized optimization.}

\subsection{Ecological Dimension: Data Imbalance and Task Dominance}
\label{subsec:data_dominance}

Finally, we examine the ecological dimension of task conflict stemming from inherent data imbalances across the training corpus. While a universal embedding model should ideally cultivate balanced representations for all semantic objectives, a monolithic adapter is inevitably hijacked by data-rich tasks. 
Figure~\ref{Fig_motivation1}(c) illustrates the layer-wise cosine similarity between the jointly trained adapter and its isolated, task-specific counterparts, exhibiting a profound ecological bias that pervades every layer.
The joint representations maintain high alignment with the data-abundant Retrieval task (peaking near 0.48), yet leave the data-scarce Visual Grounding task severely underrepresented (hovering near 0.10).

This severe representational gap demonstrates that dominant tasks unavoidably co-opt the shared parameter space throughout the entire feature extraction hierarchy. Monolithic adapters inherently fail to equitably balance heterogeneous objectives. Consequently, the model's limited capacity is monopolized by data-abundant tasks, marginalizing minority objectives and thereby undermining the core premise of truly scalable and universal multimodal representation learning.


\vspace{1mm}
\noindent\textit{\textbf{Takeaway III:} Joint training facilitates data-abundant tasks to hijack the optimization process, suppressing the representation learning of data-scarce ones.}

\section{\our: Task Scaling Multimodal Embeddings}
\label{sec:method}

In this section, we present \our, a unified multimodal embedding framework architected to establish the foundation for task-level scaling in universal representation learning.


\vspace{-1mm}
\subsection{Preliminary}\label{sec_preliminary}

\noindent\textbf{MLLM-based Embedding.}
Given an input $\mathbf{x}$, the representation is obtained by extracting the hidden state of the last token at the final layer:
\begin{equation}
    \mathbf{h} = f_{\theta}(\mathbf{x})_{[\texttt{EOS}]} \in \mathbb{R}^{d},
\end{equation}
where $f_{\theta}$ denotes the MLLM backbone and $[\texttt{EOS}]$ indexes the final token. For a query $\mathbf{q}$ and a positive target $\mathbf{k}^{+}$, the model is fine-tuned via the InfoNCE~\citep{oord2018representation} contrastive objective over a batch of $M$ negatives $\{\mathbf{k}_i^{-}\}_{i=1}^{M}$:
\begin{equation}
\label{eq:infonce}
    \mathcal{L}_{\text{InfoNCE}} = -\log \frac{\exp(s^{+} / \tau)}{\exp(s^{+} / \tau) + \sum_{i=1}^{M} \exp(s_i^{-} / \tau)},
\end{equation}
where $s^{+} = \mathbf{h}_q \cdot \mathbf{h}_{k^+}$ and $s_i^{-} = \mathbf{h}_q \cdot \mathbf{h}_{k_i^-}$ denote the query-positive and query-negative similarities, respectively, and $\tau$ is a temperature hyperparameter.

\vspace{1mm}
\noindent\textbf{Low-Rank Adaptation (LoRA).} To efficiently adapt the backbone, LoRA~\citep{hu2022lora,wu2025learning} freezes the pre-trained weights $\mathbf{W}_0 \in \mathbb{R}^{d \times k}$ and introduces a trainable low-rank residual parameterized by $\mathbf{A} \in \mathbb{R}^{r \times k}$ and $\mathbf{B} \in \mathbb{R}^{d \times r}$ ($r \ll \min(d, k)$). The adapted forward pass is defined as: $\mathbf{h'} = \mathbf{W}_0 \mathbf{x} + \mathbf{B}\mathbf{A}\mathbf{x}.$
While parameter-efficient, standard LoRA applies a \emph{uniform} transformation $\mathbf{B}\mathbf{A}$ to all inputs regardless of their semantic type~\citep{wu2025memory}, fundamentally inducing severe task conflict.

\begin{figure}[!t]
  \centering
  \includegraphics[width=1\linewidth]{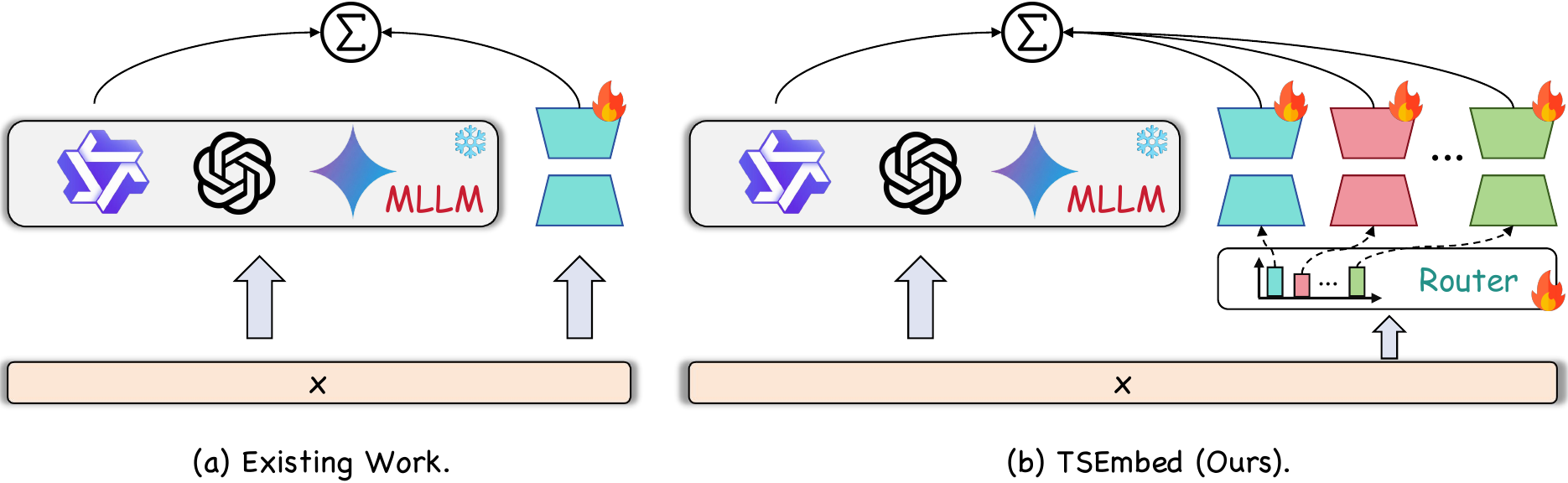}
  \caption{Overview of existing work and our proposed \our.}
  \label{fig_paradigm}
\end{figure}

\subsection{Conflict Decoupling: MoE-LoRA}\label{sec_MoE-lora}

To dismantle the monolithic parameter space that drives task conflict, we introduce the MoE-LoRA for universal multimodal embeddings (Figure~\ref{fig_paradigm}). Rather than forcing a single projection matrix to reconcile divergent tasks, this design introduces conditional computation to decouple the semantic space~\citep{wu2025elastic}. 
By routing input queries to specialized experts, it allows distinct semantic facets to be cultivated without mutual interference.
Specifically, for each layer, the adapted forward pass is reformulated as:
\begin{equation}
    \mathbf{h'} = \mathbf{W}_0 \mathbf{x} + \sum_{i=1}^{N} g_i(\mathbf{x}) \cdot \mathbf{B}_i \mathbf{A}_i \mathbf{x},
\end{equation}
where $N$ is the number of experts, and $g_i(\mathbf{x})$ denotes the routing weight for the $i$-th expert.
Formally, the gating network employs a linear projection parameterized by a weight matrix $\mathbf{W}_g \in \mathbb{R}^{N \times d}$ to dynamically compute input-dependent weights for each expert. These weights are subsequently normalized via softmax:
\begin{equation}
    \mathbf{g}(\mathbf{x}) = \text{Softmax}\left(\frac{\mathbf{W}_g \mathbf{x}}{\tau'}\right), \quad g_i(\mathbf{x}) = [\mathbf{g}(\mathbf{x})]_i,
\end{equation}
where $\tau'$ controls the sharpness of the routing distribution, and $[\cdot]_i$ indexes the $i$-th component of the resulting weight vector.

\subsection{Boundary Refinement: Expert-Aware Negative Sampling}
\label{sec_eans}

Standard contrastive learning~\citep{chen2020simple,radford2021learning} treats all negatives equally, which is fundamentally suboptimal. \emph{Hard negatives} (semantically similar to the query but with subtle differences) offer much richer gradient signals than distant \emph{trivial negatives}.
Existing hard negative mining methods~\citep{xue2025improve} typically rely on computationally prohibitive metrics or auxiliary models. In contrast, we observe that the MoE routing mechanism naturally encodes task-level semantic topology, providing a highly effective, zero-overhead proxy for identifying hard negatives.

\vspace{1mm}
\noindent\textbf{Routing Distribution as Semantic Proxy.}
For each sample processed by MoE-LoRA, we extract its routing distributions across all $L$ layers. Given that the MoE-LoRA adaptation is applied to $G$ distinct projection matrices (e.g., \texttt{Query}, \texttt{Key}, and \texttt{Value}) within each transformer block, we concatenate the routing probabilities from all $G$ matrices at layer $\ell$ to form a joint layer-wise distribution $\mathbf{g}^{(\ell)} \in [0,1]^{G \times N}$. We then aggregate these layer-wise distributions across all $L$ layers to construct a comprehensive global routing signature:
\begin{equation}
    \mathbf{r} = [\mathbf{g}^{(1)}, \mathbf{g}^{(2)}, \ldots, \mathbf{g}^{(L)}] \in [0,1]^{L \times G \times N}.
\end{equation}
The flattened vector $\text{vec}(\mathbf{r})$ comprehensively encapsulates the sample's activation trajectory throughout the network depth, explicitly mapping its task-specific expert utilization.
Given a query $\mathbf{q}$ and a negative sample $\mathbf{k}_i^-$ with their routing signatures $\mathbf{r}_q$ and $\mathbf{r}_i$, we quantify their semantic divergence as:
\begin{equation}
    d(\mathbf{r}_q, \mathbf{r}_i) = \frac{\|\text{vec}(\mathbf{r}_q) - \text{vec}(\mathbf{r}_i)\|_1}{L \cdot G \cdot N}.
\end{equation}
Normalization by the total dimension $L \cdot G \cdot N$ ensures the distance metric is scale-invariant and strictly bounded. A smaller distance indicates highly overlapping expert activation patterns, strongly suggesting that the samples share task-level semantics and thereby serve as exceptionally informative hard negatives.

\vspace{1mm}
\noindent\textbf{Exponential Decay Weighting.}
To translate the semantic distance into an effective learning signal, we design an exponential decay weighting function, which emphasizes hard negatives while smoothly suppressing trivial ones:
\begin{equation}
\label{eq:exp_decay}
    w_i = w_{\min} + (w_{\max} - w_{\min}) \cdot \exp\left(-\frac{d(\mathbf{r}_q, \mathbf{r}_i)}{\sigma}\right),
\end{equation}
where $w_{\min}$ and $w_{\max}$ establish the lower and upper bounds of the penalty, respectively, and $\sigma$ serves as a scale parameter governing the decay sensitivity. 
Compared to linear or inverse mappings, this exponential formulation exhibits superior sensitivity to minute distance variations within the high-similarity regime.
Specifically, as the semantic distance approaches zero ($d \to 0$), the weight tightly converges to $w_{\max}$, imposing a stringent penalty on hard negatives. Conversely, for semantically distant samples ($d \gg 0$), the weight rapidly decays towards $w_{\min}$, safely marginalizing trivial negatives.

\vspace{1mm}
\noindent\textbf{Weight Normalization for Gradient Consistency.}
To ensure that the dynamic weighting mechanism does not inadvertently distort the global gradient scale relative to the standard InfoNCE loss, we normalize the raw weights such that their sum strictly equals the total number of negative samples, $M$:
\begin{equation}
\label{eq:weight_norm}
    \tilde{w}_i = w_i \cdot \frac{M}{\sum_{j=1}^{M} w_j}.
\end{equation}
This ensures the aggregate negative contribution in the loss denominator remains strictly invariant. As a result, EANS purely redistributes the gradient budget, concentrating optimization capacity on informative hard negatives without compromising the overall stability of the training dynamics.


\vspace{1mm}
\noindent\textbf{EANS Loss.}
The final Expert-Aware Negative Sampling loss is:
\begin{equation}
\label{eq:eans_loss}
    \mathcal{L}_{\text{EANS}} = -\log \frac{\exp(s^+ / \tau)}{\exp(s^+ / \tau) + \sum_{i=1}^{M} \textcolor{red}{\tilde{w}_i} \cdot \exp(s_i^- / \tau)}.
\end{equation}
EANS effectively compels hard negatives with high $\tilde{w}_i$ to exert a greater influence, thereby sharpening the model’s representational capability.

\subsection{Two-Stage Learning Paradigm}
\label{subsec:training}

The efficacy of EANS hinges on a critical prerequisite: routing distributions must serve as reliable semantic proxies. 
Since randomly initialized routers are inherently stochastic, applying EANS prematurely would inject spurious gradients and destabilize training. Therefore, we design a progressive two-stage paradigm.

\vspace{1mm}
\noindent\textbf{Stage 1: Expert Warm-up ($t < T_{\text{warmup}}$).}
Optimized exclusively via standard InfoNCE, this phase allows MoE-LoRA to naturally disentangle the heterogeneous semantic space.
Guided purely by data distributions, experts autonomously carve out distinct functional niches to resolve task conflicts.

\vspace{1mm}
\noindent\textbf{Stage 2: EANS Refinement ($t \geq T_{\text{warmup}}$).}
Upon the stabilization of the routing topology, we engage the dynamically weighted EANS loss. Leveraging the reliable routing trajectories as semantic cues, EANS imposes targeted penalties on hard negatives, effectively sharpening the embedding boundaries.

\vspace{1mm}
\noindent\textbf{Overall Objective.}
Formally, we optimize the piecewise combined loss:
\begin{equation}
    \mathcal{L} = \begin{cases}
        \mathcal{L}_{\text{InfoNCE}}  & \text{if } t < T_{\text{warmup}}, \\
        \mathcal{L}_{\text{EANS}} & \text{if } t \geq T_{\text{warmup}}.
    \end{cases}
\end{equation}

\section{Experiments}
\label{sec:experiments}

\subsection{Experimental Setup}

\noindent\textbf{Datasets and Models.} We evaluate \our on the MMEB~\citep{jiang2024vlm2vec} alongside proprietary real-world industrial datasets to rigorously assess both in-domain alignment and zero-shot generalization capabilities. Qwen2-VL~\citep{wang2024qwen2} is employed as the backbone, with evaluations scaled across its 2B and 7B variants.

\vspace{1mm}
\noindent\textbf{Implementation Details.} We implement \our in PyTorch, executing all experiments on 8 NVIDIA A800 GPUs. During visual preprocessing, images are resized with a minimum resolution of 401,408 pixels. We extract the sequence embedding from the final hidden state corresponding to the \texttt{<EOS>} token. To facilitate large-batch contrastive learning within hardware memory constraints, we integrate Gradient Caching~\citep{gao2021scaling}, configuring a sub-batch chunk size of 2.

\vspace{1mm}
\noindent\textbf{Hyperparameter Settings.}
We optimize the model using AdamW with a learning rate of $5 \times 10^{-5}$ and a linear decay scheduler.
The training process spans 2,200 steps with a global batch size of 1,024 (128 per GPU).
Moreover, our MoE layers feature $N=4$ experts, with each LoRA configured with a rank of 16 and a scaling factor of 64.
The exponential decay weighting function is parameterized by  $w_{\min}=0.1$, $w_{\max}=10.0$, and a decay sensitivity $\sigma=0.002$. $T_{\text{warmup}}$ is set to 600 and 1,200 steps for the 2B and 7B models, respectively.

\subsection{Baselines}

We compare \our against a wide range of multimodal embedding models:
\begin{itemize}
    \item \textbf{Encoder-Only Models:} We include representative dual-encoder architectures such as CLIP~\citep{radford2021learning}, OpenCLIP~\citep{cherti2023reproducible}, SigLIP~\citep{tschannen2025siglip}, and BLIP-2~\citep{li2023blip}, as well as retrieval-specialized models like UniIR~\citep{wei2024uniir} and Magiclens~\citep{zhang2024magiclens}.

    \item \textbf{MLLM-Based Embedding Models:} We benchmark against state-of-the-art generative embedding models across various scales, including VLM2VEC~\citep{jiang2024vlm2vec}, UniME-V2~\citep{gu2025unime}, LLaVE~\citep{lan2025llave}, B3~\citep{thirukovalluru2025breaking}, and QQMM-embed~\citep{xue2025improve}.

    \item \textbf{External Data-Augmented Models:} We also include the performance of the proprietary Seed-1.6-embedding model as a commercial benchmark, alongside a suite of models augmented with massive external corpora beyond MMEB. These include Ops-MM-embedding-v1, GME~\citep{zhang2025bridging}, UNITE~\citep{kong2025modality}, CAFe~\citep{yu2025cafe}, ColPali-v1.3~\citep{faysse2024colpali}, LamRA~\citep{liu2025lamra}, RzenEmbed~\citep{jian2025rzenembed}, and Qwen3-VL-Embedding~\citep{li2026qwen3}. It is important to note that these models serve solely as high-resource references rather than direct competitors. To ensure a fair assessment of algorithmic contributions, our primary evaluation focuses on models trained exclusively on the MMEB dataset, thereby isolating the architectural efficacy of our proposed framework.


\end{itemize}

\begin{table*}[!t]
	\centering
    \renewcommand{\arraystretch}{1.1}
    \caption{\textcolor{darkred}{Performance comparison of various multimodal embedding models on the MMEB~\citep{jiang2024vlm2vec}.} \textbf{Bold} and \underline{underlined} values denote the best and second-best results within each parameter scale, respectively. $\textsuperscript{\dag}$ indicates a link to the model's official homepage.}
	\label{tab:main}
	\resizebox{\textwidth}{!}{
		\begin{tabular}{lccccccccccc}
			\toprule
			\multirow{2}{*}{\textbf{Model}} & \multirow{2}{*}{\textbf{Backbone}} & \multirow{2}{*}{\textbf{Model Size}} & \multicolumn{4}{c}{\textbf{Per Meta-Task Score}} & & \multicolumn{3}{c}{\textbf{Average Score}} \\ 
			\cmidrule(lr){4-7} \cmidrule(lr){9-11}
			& & & \textbf{Classification} & \textbf{VQA}  & \textbf{Retrieval} & \textbf{Grounding} & & \textbf{IND} & \textbf{OOD} & \textbf{Overall} \\ \midrule
			\textbf{\# of datasets} $\rightarrow$ & & & 10 & 10 & 12 & 4 & & 20 & 16 & 36 \\ \midrule
			
			\multicolumn{11}{c}{\textbf{\emph{Encoder-Only Models}}} \\ \midrule
			\rowcolor{gray!10}
			CLIP~\citep{radford2021learning}   & -  & 0.428B  & 42.8 & 9.1 &  53.0 &  51.8 &   &  37.1  &  38.7 &  37.8 \\
			\rowcolor{white}
			BLIP-2~\citep{li2023blip}    & - & 3.74B & 27.0  &  4.2 & 33.9  & 47.0 &  &  25.3 &  25.1 & 25.2 \\
			\rowcolor{gray!10}
			SigLIP~\citep{tschannen2025siglip}   &  - & 0.203B & 40.3  &  8.4 & 31.6  & 59.5 &  &  32.3 &  38.0 & 34.8 \\
			\rowcolor{white}
			OpenCLIP~\citep{cherti2023reproducible}   & -  & 0.428B & 47.8  &  10.9 & 52.3  & 53.3 &  &  39.3 &  40.2 & 39.7 \\
			\rowcolor{gray!10}
			UniIR (BLIP\_FF)~\citep{wei2024uniir}   & -  & 0.247B &  42.1 &	 15.0  &	60.1 & 	62.2  &	 & 44.7	&  40.4 & 	42.8 \\
			\rowcolor{white}
			UniIR (CLIP\_SF)~\citep{wei2024uniir}   & -  & 0.428B & 44.3 & 16.2 & 61.8 & 65.3 & & 47.1 & 41.7 & 44.7 \\
			\rowcolor{gray!10}
			Magiclens~\citep{zhang2024magiclens}   & -  & 0.428B &  38.8 &  8.3  &  35.4 &  26.0 &  & 31.0 & 23.7  & 27.8  \\
            \midrule
            \multicolumn{11}{c}{\textbf{\emph{External Data-Augmented Models}}} \\ \midrule
            \rowcolor{gray!10}
            Seed-1.6-embedding\href{https://seed1-6-embedding.github.io/}{$\textsuperscript{\dag}$} & Seed1.6-flash  & unknown & 76.1 & 74.0 & 77.9 & 91.3 &  & -  & - & 77.8 \\
            Ops-MM-embedding-v1\href{https://huggingface.co/OpenSearch-AI/Ops-MM-embedding-v1-2B}{$\textsuperscript{\dag}$}   & Qwen2-VL &  2.21B & 68.1 & 65.1 & 69.2 & 80.9 &   &  - &-  & 69.0  \\
            \rowcolor{gray!10}
			Ops-MM-embedding-v1\href{https://huggingface.co/OpenSearch-AI/Ops-MM-embedding-v1-7B}{$\textsuperscript{\dag}$}     & Qwen2-VL &  8.29B & 69.7 & 69.6  & 73.1 & 87.2 &   &  - &-  & 72.7  \\

			GME~\citep{zhang2025bridging}    & Qwen2-VL  & 2.21B  & 54.4 & 29.9 &66.9  &55.5  &   &   - & -  & 51.9 \\
            \rowcolor{gray!10}
			GME~\citep{zhang2025bridging}   & Qwen2-VL  & 8.29B  & 57.7 & 34.7 & 71.2 & 59.3 &   &  - & - &56.0  \\
            UNITE~\citep{kong2025modality}     & Qwen2-VL  &  2.21B &63.2  &55.9  & 65.4 &75.6  &   &65.8   & 60.1 & 63.3 \\
            \rowcolor{gray!10}
            UNITE~\citep{kong2025modality}    & Qwen2-VL  & 8.29B  &68.3  & 65.1 & 71.6 & 84.8 &   & 73.6  & 66.3 &70.3  \\
            CAFe~\citep{yu2025cafe} & LLaVA-OV & 0.894B & 59.1 & 49.1 & 61.0 & 83.0 &   & 64.3  & 53.7 & 59.6 \\
            \rowcolor{gray!10}
            CAFe~\citep{yu2025cafe} & LLaVA-OV & 8.03B & 65.2 &65.6 & 70.0 & 91.2 &   & 75.8   & 62.4 & 69.8  \\
			ColPali-v1.3~\citep{faysse2024colpali}    & PaliGemma &  2.92B &40.3  & 11.5 &48.1  & 40.3 &   &-   & - &34.9  \\
            \rowcolor{gray!10}
            LamRA~\citep{liu2025lamra}  & Qwen2-VL  & 8.29B  & 59.2 & 26.5 &  70.0& 62.7 &   & -  & - & 54.1 \\
			
			LamRA~\citep{liu2025lamra} & Qwen2.5-VL  & 8.29B  & 51.7 & 34.1 &66.9 & 56.7 &   &  -  & -  & 52.4 \\
            \rowcolor{gray!10}
			RzenEmbed~\citep{jian2025rzenembed} & Qwen2-VL  & 2.21B  & 68.5  & 66.3 & 74.5 & 90.3 &  & 76.1   & 67.4 & 72.3 \\
            RzenEmbed~\citep{jian2025rzenembed}  & Qwen2-VL  & 8.29B   & 70.6 & 71.7 & 78.5 & 92.1 &   & 78.5  & 72.7 & 75.9 \\
            \rowcolor{gray!10}
            Qwen3-VL-Embedding~\citep{li2026qwen3} & Qwen3-VL & 2.13B &  70.3 & 74.3 &  74.8 & 88.5 & & 75.3 & 74.5 & 75.0\\
            Qwen3-VL-Embedding~\citep{li2026qwen3} & Qwen3-VL & 8.14B &  74.2 & 81.1 &  80.2 & 92.3 & & 81.9 & 78.0 & 80.1\\
            
			\midrule

			\multicolumn{11}{c}{\textbf{\emph{$\sim$ 2B Models}}} \\ \midrule
			\rowcolor{gray!10}
			VLM2VEC~\citep{jiang2024vlm2vec} & Phi-3.5-V  & 4.15B  &54.8  &54.9  &62.3  &79.5  &   & 66.5  &52.0  & 60.1  \\
			VLM2VEC~\citep{jiang2024vlm2vec}  & Qwen2-VL  & 2.21B  & 59.0 & 49.4 &  65.4& 73.4 &   & 66.0  & 52.6 & 59.3 \\
			\rowcolor{gray!10}
			VLM2VEC-V2~\citep{meng2025vlm2vec}   & Qwen2-VL  & 2.21B  &62.9  &56.3  &69.5  &77.3  &   &  - & - &64.9  \\
			UniME-V2~\citep{gu2025unime}   & Qwen2-VL  & 2.21B  & 62.1 & 56.3 & 68.0 & 72.7 &   &67.4   &58.9  &63.6  \\
            \rowcolor{gray!10}
			LLaVE~\citep{lan2025llave}   & Aquila-VL  & 1.95B  & 62.1 & 60.2 &65.2  &\underline{84.9}  &   & 69.4  & 59.8 &65.2  \\
			B3~\citep{thirukovalluru2025breaking}   & Qwen2-VL  &  2.21B & \underline{67.0}  & \underline{61.2} & \underline{70.9} & 79.9 &   & \underline{72.1}  &\underline{63.1}  & \underline{68.1} \\
			
			\rowcolor{blue!5}
    			\textbf{\our (ours)}   & Qwen2-VL  & 2.26B &  \textbf{68.8} & \textbf{64.3} & \textbf{72.1} & \textbf{85.7} &  & \textbf{74.5} & \textbf{65.6} & \textbf{70.5} \\

			\midrule	
			\multicolumn{11}{c}{\textbf{\emph{$\sim$ 7B Models}}} \\ \midrule
			\rowcolor{gray!10}
			VLM2VEC~\citep{jiang2024vlm2vec}   & LLaVA-1.6  & 7.57B  & 61.2 &49.9 & 67.4 &86.1  &   & 67.5  & 57.1 & 62.9 \\
			VLM2VEC~\citep{jiang2024vlm2vec}   & Qwen2-VL  & 8.29B  & 62.6 & 57.8 & 69.9 & 81.7 &   & 65.2  & 56.3 & 65.8 \\
			
			\rowcolor{gray!10}
			UniME-V2~\citep{gu2025unime}    & LLaVA-OV  & 8.03B  & 65.3 & \underline{67.6} & 72.9 & 90.2 &   & 74.8  & 66.7 & 71.2 \\
			UniME-V2~\citep{gu2025unime}    & Qwen2-VL  & 8.29B  & 64.0 &60.1  & 73.1 & 82.8 &   & 72.0  &63.0  &68.0  \\	
            \rowcolor{gray!10}
			LLaVE~\citep{lan2025llave}    & LLaVA-OV  & 8.03B  & 65.7 & 65.4 &  70.9&  \textbf{91.9} &  &  75.0 & 64.4 & 70.3 \\
			
			B3~\citep{thirukovalluru2025breaking}   & Qwen2-VL  &  8.29B & \underline{70.0} &66.5  & \underline{74.1} & 84.6 &   &\underline{75.9}   &\underline{67.1}  & \underline{72.0} \\

			\rowcolor{gray!10}
			QQMM-embed~\citep{xue2025improve}  & LLaVA-OV  &  8.297B & 66.8 & 66.8 & 70.5 &  90.4 &   &  74.7 & 65.6 & 70.7 \\

			\rowcolor{blue!5}
			\textbf{\our (ours)}  & Qwen2-VL  & 8.40B   & \textbf{71.1} & \textbf{70.3} & \textbf{75.9} & \underline{91.3} & & \textbf{78.8} & \textbf{69.6} &  \textbf{74.7} \\
			
			\bottomrule
		\end{tabular}
	}
\end{table*}


\subsection{Main Results}

Table~\ref{tab:main} summarizes the comparative results on the MMEB, where \our demonstrates consistent superiority over baselines across all evaluation settings.

\vspace{1mm}
\noindent\textbf{Overall Performance.}
Among MLLM-based models trained exclusively on the standard MMEB dataset, \our achieves new state-of-the-art performance with \textbf{74.7\%} at the 7B scale and \textbf{70.5\%} at the 2B scale.
At the 7B scale, \our surpasses the previous best method, B3 (72.0\%), by a margin of \textbf{2.7\%}, while consistently outperforming other strong baselines such as UniME-V2 (+3.5\%), QQMM-embed (+4.0\%), and LLaVE (+4.4\%).
This superiority extends to the 2B scale, where \our improves over B3 (+2.4\%) and LLaVE (+5.3\%).
Most notably, compared to the representative VLM2VEC (65.8\% at 7B), \our yields a substantial improvement of \textbf{8.9\%}.
Crucially, \our achieves superior performance compared to several baselines augmented with external data, despite being constrained strictly to the MMEB training set: it outperforms Ops-MM-embedding-v1 by \textbf{2.0\%} at the 7B scale (74.7\% vs.\ 72.7\%) and by \textbf{1.5\%} at the 2B scale (70.5\% vs.\ 69.0\%), and substantially exceeds data-augmented models such as UNITE (74.7\% vs.\ 70.3\%) and CAFe (74.7\% vs.\ 69.8\%) at the 7B scale. 
These results demonstrate that our principled architectural design not only mitigates task conflict, but also unlocks exceptional data efficiency, effectively bridging the performance gap with models trained on external corpora.


\vspace{1mm}
\noindent\textbf{Approaching Task-Specific Oracle Performance.}
Notably, \our achieves strong per-task performance: at the 7B scale, it attains 71.1\% on Classification, 70.3\% on VQA, 75.9\% on Retrieval, and 91.3\% on Visual Grounding, closely approaching or even surpassing the oracle performance set by dedicated task-specific models (71.5\%, 70.9\%, 75.5\%, and 91.7\%, respectively; Figure~\ref{fig:motivation}). 
These results collectively demonstrate that \our effectively preserves task-specific specialization within a unified framework, validating the task-scaling capability of our MoE-LoRA for universal multimodal embeddings.


\vspace{1mm}
\noindent\textbf{Robustness Across Data Distributions.} Beyond merely achieving strong in-domain alignment, \our demonstrates an exceptional zero-shot generalization capability. At the 7B scale, the model attains 78.8\% on in-distribution (IND) and 69.6\% on out-of-distribution (OOD) tasks, outperforming the formidable B3 baseline by 2.9\% and 2.5\%, respectively. This robust performance seamlessly extends to the 2B scale, where \our achieves 74.5\% IND and 65.6\% OOD, surpassing B3 by 2.4\% and 2.5\%. These consistent performance gains across varying model scales and data regimes confirm that our MoE-driven decomposition encourages experts to cultivate intrinsically transferable semantic patterns, rather than overfitting to the training distribution.

\begin{figure}[t!]
    \begin{minipage}{\textwidth}

    \begin{minipage}[t]{0.38\linewidth}
    \vspace{0pt}
    \captionof{table}{Performance evaluation on \textcolor{darkred}{proprietary production datasets} sourced from a large-scale technology enterprise.}
    \label{tab:oppo_results}
    \renewcommand{\arraystretch}{1.1}
    \resizebox{\linewidth}{!}{
    \begin{tabular}{c|c|c|c}
    \hline\hline
    Domain & Metric & VLM2VEC & \our \\
    \hline\hline
    Advertising & Recall  & 11.33\% & \textbf{33.20\%} {\color{green!50!black}\scriptsize($\uparrow$ 21.87)} \\
    Theme      & NDCG@5  & 84.17\% & \textbf{86.22\%} {\color{green!50!black}\scriptsize($\uparrow$ 2.05)}  \\
    Lockscreen  & NDCG@1  & 56.31\% & \textbf{61.54\%} {\color{green!50!black}\scriptsize($\uparrow$ 5.23)}  \\
    Gaming      & NDCG@1  & 36.07\% & \textbf{40.06\%} {\color{green!50!black}\scriptsize($\uparrow$ 3.99)}  \\
    \hline\hline
    \end{tabular}}
    \end{minipage}
    \hfill
    \begin{minipage}[t]{0.6\linewidth}
    \vspace{0pt}
    \definecolor{vlm2vec_color}{RGB}{252,146,114}
    \definecolor{our_color}{RGB}{46,134,171}
    \centering
    \begin{tikzpicture}
        \tiny{
        \begin{axis}[
            anchor=north west,
            at={(0em,0em)},
            ymajorgrids,
            grid style=dashed,
            ybar=3pt,
            enlarge x limits=0.65,
            xtick align=inside,
            width=.6\textwidth,
            height=.4\textwidth,
            bar width=1.5em,
            ymin=0, ymax=120,
            ytick={0,30,60,90,120},
            symbolic x coords={{1},{2}},
            xtick=data,
            xticklabels={2B, 7B},
            xticklabel style={scale=1.1, yshift=-0.3em},
            ylabel={\scriptsize{Time (h)}},
            ylabel style={yshift=-3em, scale=1.15},
            yticklabel style={/pgf/number format/precision=1,/pgf/number format/fixed,scale=0.95},
            xlabel={\footnotesize(a) Training Efficiency.},
            xlabel style={yshift=0.3em, scale=1.1},
            legend style={at={(1.96,1.5)}, anchor=north east, font={\scriptsize}, cells={anchor=west}, fill opacity=0.9, scale=0.8, legend columns=2, column sep=1em}
            ]
            \addplot[fill=vlm2vec_color, draw=black, line width=0.8pt, area legend]
                coordinates {({1},45.89) ({2},63.42)};
            \addlegendentry{\scalebox{1.0}{VLM2VEC}}
            \addplot[fill=our_color, draw=black, line width=0.8pt, area legend,
                     nodes near coords={+\pgfmathprintnumber[fixed,precision=2]{\pgfplotspointmeta}},
                     point meta=explicit,
                     every node near coord/.style={anchor=south, font=\tiny, yshift=0.1em}]
                coordinates {({1},65.42)[19.53] ({2},83.47)[20.05]};
            \addlegendentry{\scalebox{1.0}{\our}}
        \end{axis}
        \begin{axis}[
            anchor=north west,
            at={(15em,0em)},
            ymajorgrids,
            grid style=dashed,
            ybar=3pt,
            enlarge x limits=0.65,
            xtick align=inside,
            width=.6\textwidth,
            height=.4\textwidth,
            bar width=1.5em,
            ymin=0, ymax=12,
            ytick={0,4,8,12},
            symbolic x coords={{1},{2}},
            xtick=data,
            xticklabels={2B, 7B},
            xticklabel style={scale=1.1, yshift=-0.3em},
            ylabel={\scriptsize{Para. (B)}},
            ylabel style={yshift=-3em, scale=1.15},
            yticklabel style={/pgf/number format/precision=1,/pgf/number format/fixed,scale=0.95},
            xlabel={\footnotesize (b) Model Size.},
            xlabel style={yshift=0.3em, scale=1.1}
            ]
            \addplot[fill=vlm2vec_color, draw=black, line width=0.8pt, area legend]
                coordinates {({1},2.219) ({2},8.311)};
            \addplot[fill=our_color, draw=black, line width=0.8pt, area legend,
                     nodes near coords={+\pgfmathprintnumber[fixed,precision=3]{\pgfplotspointmeta}},
                     point meta=explicit,
                     every node near coord/.style={anchor=south, font=\tiny, yshift=0.1em}]
                coordinates {({1},2.257)[0.038] ({2},8.395)[0.084]};
        \end{axis}
        }
    \end{tikzpicture}
    \captionof{figure}{\textcolor{darkred}{Efficiency analysis} of \our.}
    \label{fig:efficiency}
    \end{minipage}

    \end{minipage}
\vspace{-3mm}
\end{figure}
\vspace{-2mm}
\subsection{Generalization and Efficiency Analysis}

\noindent\textbf{Generalization Analysis.} To evaluate its real-world efficacy, we validate \our on proprietary production datasets encompassing diverse commercial domains, including advertising, theme, lockscreen, and gaming. 
\our consistently exhibits robust zero-shot generalization without requiring any domain-specific fine-tuning, surpassing VLM2VEC by a significant margin (Table~\ref{tab:oppo_results}).
Specifically, it achieves a substantial 21.87\% gain in advertising, alongside steady improvements in theme (+2.05\%), lockscreen (+5.23\%), and gaming (+3.99\%).
These results substantiate that \our effectively learns transferable multimodal representations rather than merely memorizing patterns from academic datasets.

\vspace{1mm}
\noindent\textbf{Training Efficiency.} 
Figure~\ref{fig:efficiency}(a) demonstrates that \our introduces minimal training overhead, incurring only an additional 19.53 and 20.05 hours of training time for the 2B and 7B models, respectively. This slight increase is primarily attributed to
the computations required for expert routing and EANS. 
Given its computational efficiency and stable scaling behavior, \our is conducive to large-scale industrial deployment.

\vspace{1mm}
\noindent\textbf{Parameter Efficiency.} Furthermore, \our demonstrates exceptional structural scalability. As shown in Figure~\ref{fig:efficiency}(b), it requires a mere 0.038B (+1.7\%) and 0.084B (+1.0\%) additional parameters for the 2B and 7B models, respectively. 
Given the significant performance gain of up to 11.8\% over VLM2VEC at the 7B scale, \our delivers an optimal capacity-to-performance trade-off, effectively amplifying representational power with negligible parameter expansion.

\subsection{Sensitivity Analysis}

\noindent\textbf{EANS Decay Sensitivity ($\sigma$).}
The decay parameter $\sigma$ (Eq.~\ref{eq:exp_decay}) governs hard-negative weighting sharpness. Remarkably, \our demonstrates exceptional stability across a vast range of $\sigma \in [2\!\times\!10^{-5}, 2\!\times\!10^{-1}]$ (Figure~\ref{fig:sensitivity_theta}). Across all meta-tasks and model scales, performance fluctuations are tightly bounded within a ${\sim}1\%$ margin. 
Although extreme values induce slight degradations, stemming from weight homogenization at large $\sigma$ and gradient over-concentration at small $\sigma$, the overall performance remains remarkably resilient. This striking consistency validates the inherent robustness of our framework.

\definecolor{red}{RGB}{172,21,28}
\definecolor{blue}{RGB}{39,89,167}
\definecolor{red1}{RGB}{203,104,104}
\definecolor{blue1}{RGB}{104,155,203}
\definecolor{color1}{HTML}{283c63}
\definecolor{color2}{HTML}{00ad7c}

\begin{figure*}[!t]
\centering
\hspace{-2mm}
\begin{tikzpicture}
    \tiny{
    \begin{axis}
    [
        anchor=north west,
        at={(-40em,-5em)},
        ymajorgrids,
        xmajorgrids,
        grid style=dashed,
        width=.31\textwidth,
        height=.26\textwidth,
        yticklabel style={/pgf/number format/precision=0,/pgf/number format/fixed zerofill,scale=1.0},
        xmax=2100,
        xmin=300,
        ymin=67,
        ymax=72.5,
        xtick={400,800,1200,1600,2000},
        xticklabels={2e-5,2e-4,2e-3,2e-2,2e-1},
        ytick={67,69,71},
        xlabel={\scriptsize{(a) Classification.}},
        xlabel style={scale=1.2, yshift=0.2em, xshift=0.1em},
        ylabel={\scriptsize hit@1 (\%)},
        ylabel style={yshift=-3.em, scale=1.2},
        legend style={at={(3.5,1.4)}, anchor=north east, font={\tiny}, cells={anchor=west}, fill opacity=0.8, scale=1.2, legend columns=3}
        ]

        \addplot[red,mark=pentagon*,,mark size=2.5pt,thick,mark options={fill=white,draw=red,line width=1pt}] coordinates {(400,68.00) (800,68.32) (1200,68.77) (1600,68.32) (2000,67.91)};
        \addlegendentry{\scalebox{1.2}{\our-2B}}

        \addplot[color1,mark=*,mark size=2.5pt,thick,mark options={fill=white,draw=color1,line width=1pt}] coordinates {(400,71.34) (800,71.01) (1200,71.07) (1600,71.28) (2000,70.93)};
        \addlegendentry{\scalebox{1.2}{\our-7B}}

    \end{axis}
	
  \begin{axis}
    [
        anchor=north west,
        at={(-27.5em,-5em)},
        ymajorgrids,
        xmajorgrids,
        grid style=dashed,
        width=.31\textwidth,
        height=.26\textwidth,
        yticklabel style={/pgf/number format/precision=0,/pgf/number format/fixed zerofill,scale=1.0},
        xmax=2100,
        xmin=300,
        ymin=62,
        ymax=72,
        xtick={400,800,1200,1600,2000},
        xticklabels={2e-5,2e-4,2e-3,2e-2,2e-1},
        ytick={62,66,70},
        xlabel={\scriptsize{(b) VQA.}},
        xlabel style={scale=1.2, yshift=0.2em, xshift=0.1em},
        ylabel=\footnotesize{\scriptsize hit@1 (\%)},
        ylabel style={yshift=-3.em, scale=1.2},
        legend style={at={(0.5,1.2)}, anchor=north east, font={\tiny}, cells={anchor=west}, fill opacity=0.8, scale=1.0, legend columns=3}
        ]

        \addplot[red,mark=pentagon*,,mark size=2.5pt,thick,mark options={fill=white,draw=red,line width=1pt}] coordinates {(400,64.32) (800,63.93) (1200,64.29) (1600,64.22) (2000,63.97)};

        \addplot[color1,mark=*,mark size=2.5pt,thick,mark options={fill=white,draw=color1,line width=1pt}] coordinates {(400,70.04) (800,70.00) (1200,70.26) (1600,70.43) (2000,70.21)};
    \end{axis}}   

    \begin{axis}
    [
        anchor=north west,
        at={(-15em,-5em)},
        ymajorgrids,
        xmajorgrids,
        grid style=dashed,
        width=.31\textwidth,
        height=.26\textwidth,
        yticklabel style={/pgf/number format/precision=0,/pgf/number format/fixed zerofill,scale=1.0},
        xmax=2100,
        xmin=300,
        ymin=70,
        ymax=77,
        xtick={400,800,1200,1600,2000},
        xticklabels={2e-5,2e-4,2e-3,2e-2,2e-1},
        ytick={70,73,76},
        xlabel={\scriptsize{(c) Retrieval.}},
        xlabel style={scale=1.2, yshift=0.2em, xshift=0.1em},
        ylabel=\footnotesize{\scriptsize hit@1 (\%)},
        ylabel style={yshift=-3.em, scale=1.2},
        legend style={at={(0.45,1.2)}, anchor=north east, font={\tiny}, cells={anchor=west}, fill opacity=0.8, scale=1.0, legend columns=3}
        ]

        \addplot[red,mark=pentagon*,,mark size=2.5pt,thick,mark options={fill=white,draw=red,line width=1pt}] coordinates {(400,72.18) (800,71.88) (1200,72.09) (1600,71.84) (2000,72.00)};

        \addplot[color1,mark=*,mark size=2.5pt,thick,mark options={fill=white,draw=color1,line width=1pt}] coordinates {(400,75.63) (800,75.95) (1200,75.85) (1600,75.41) (2000,75.70)};
    \end{axis}

    \begin{axis}
    [
        anchor=north west,
        at={(-2.5em,-5em)},
        ymajorgrids,
        xmajorgrids,
        grid style=dashed,
        width=.31\textwidth,
        height=.26\textwidth,
        yticklabel style={/pgf/number format/precision=0,/pgf/number format/fixed zerofill,scale=1.0},
        xmax=2100,
        xmin=300,
        ymin=84,
        ymax=93,
        xtick={400,800,1200,1600,2000},
        xticklabels={2e-5,2e-4,2e-3,2e-2,2e-1},
        ytick={84,88,92},
        xlabel={\scriptsize{(d) Grounding.}},
        xlabel style={scale=1.2, yshift=0.2em, xshift=0.1em},
        ylabel=\footnotesize{\scriptsize hit@1 (\%)},
        ylabel style={yshift=-3.em, scale=1.2},
        legend style={at={(0.5,1.2)}, anchor=north east, font={\tiny}, cells={anchor=west}, fill opacity=0.8, scale=1.0, legend columns=3}
        ]

        \addplot[red,mark=pentagon*,,mark size=2.5pt,thick,mark options={fill=white,draw=red,line width=1pt}] coordinates {(400,85.53) (800,85.83) (1200,85.73) (1600,85.58) (2000,85.55)};

        \addplot[color1,mark=*,mark size=2.5pt,thick,mark options={fill=white,draw=color1,line width=1pt}] coordinates {(400,90.95) (800,91.23) (1200,91.25) (1600,91.15) (2000,91.10)};
        
    \end{axis}  
    \end{tikzpicture}
    \caption{\textcolor{darkred}{Sensitivity analysis of the EANS decay parameter $\sigma$.} We evaluate the impact of $\sigma$ on four meta-task categories using both 2B and 7B models.}
    \label{fig:sensitivity_theta}
\end{figure*}
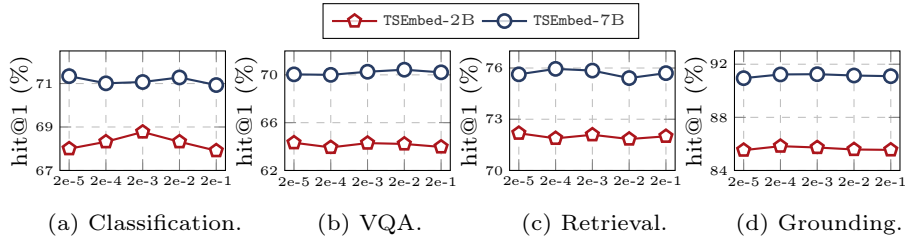
\definecolor{red}{RGB}{172,21,28}
\definecolor{blue}{RGB}{39,89,167}
\definecolor{red1}{RGB}{203,104,104}
\definecolor{blue1}{RGB}{104,155,203}
\definecolor{color1}{HTML}{283c63}
\definecolor{color2}{HTML}{00ad7c}
\definecolor{purple}{RGB}{142,68,173}
\definecolor{orange}{RGB}{230,126,34}

\begin{figure*}[!t]
\centering
\hspace{-2mm}
\begin{tikzpicture}
    \tiny{
    \begin{axis}
    [
        anchor=north west,
        at={(-40em,-5em)},
        ymajorgrids,
        xmajorgrids,
        grid style=dashed,
        width=.31\textwidth,
        height=.26\textwidth,
        yticklabel style={/pgf/number format/precision=0,/pgf/number format/fixed zerofill,scale=1.0},
        xmax=2100,
        xmin=300,
        ymin=67,
        ymax=72,
        xtick={400,800,1200,1600,2000},
        xticklabels={300,600,900,1200,1500},
        ytick={67,69,71},
        xlabel={\scriptsize{(a) Classification.}},
        xlabel style={scale=1.2, yshift=0.2em, xshift=0.1em},
        ylabel=\footnotesize{\scriptsize hit@1 (\%)},
        ylabel style={yshift=-3.em, scale=1.2},
        legend style={at={(3.5,1.4)}, anchor=north east, font={\tiny}, cells={anchor=west}, fill opacity=0.8, scale=1.0, legend columns=3}
        ]

        \addplot[purple,mark=square*,mark size=2.5pt,thick,mark options={fill=white,draw=purple,line width=1pt}] coordinates {(400,68.24) (800,68.77) (1200,68.14) (1600,68.31) (2000,68.24)};
        \addlegendentry{\scalebox{1.2}{\our-2B}}

        \addplot[orange,mark=diamond*,mark size=3pt,thick,mark options={fill=white,draw=orange,line width=1pt}] coordinates {(400,70.68) (800,70.09) (1200,71.13) (1600,71.07) (2000,70.80)};
        \addlegendentry{\scalebox{1.2}{\our-7B}}

    \end{axis}
	
  \begin{axis}
    [
        anchor=north west,
        at={(-27.5em,-5em)},
        ymajorgrids,
        xmajorgrids,
        grid style=dashed,
        width=.31\textwidth,
        height=.26\textwidth,
        yticklabel style={/pgf/number format/precision=0,/pgf/number format/fixed zerofill,scale=1.0},
        xmax=2100,
        xmin=300,
        ymin=62,
        ymax=72,
        xtick={400,800,1200,1600,2000},
        xticklabels={300,600,900,1200,1500},
        ytick={62,66,70},
        xlabel={\scriptsize{(b) VQA.}},
        xlabel style={scale=1.2, yshift=0.2em, xshift=0.1em},
        ylabel=\footnotesize{\scriptsize hit@1 (\%)},
        ylabel style={yshift=-3.em, scale=1.2},
        legend style={at={(0.5,1.2)}, anchor=north east, font={\tiny}, cells={anchor=west}, fill opacity=0.8, scale=1.0, legend columns=3}
        ]

        \addplot[purple,mark=square*,mark size=2.5pt,thick,mark options={fill=white,draw=purple,line width=1pt}] coordinates {(400,63.87) (800,64.29) (1200,63.74) (1600,63.85) (2000,64.13)};

        \addplot[orange,mark=diamond*,mark size=3pt,thick,mark options={fill=white,draw=orange,line width=1pt}] coordinates {(400,70.37) (800,70.41) (1200,70.81) (1600,70.26) (2000,69.97)};
    \end{axis}}   

    \begin{axis}
    [
        anchor=north west,
        at={(-15em,-5em)},
        ymajorgrids,
        xmajorgrids,
        grid style=dashed,
        width=.31\textwidth,
        height=.26\textwidth,
        yticklabel style={/pgf/number format/precision=0,/pgf/number format/fixed zerofill,scale=1.0},
        xmax=2100,
        xmin=300,
        ymin=70,
        ymax=77,
        xtick={400,800,1200,1600,2000},
        xticklabels={300,600,900,1200,1500},
        ytick={70,73,76},
        xlabel={\scriptsize{(c) Retrieval.}},
        xlabel style={scale=1.2, yshift=0.2em, xshift=0.1em},
        ylabel=\footnotesize{\scriptsize hit@1 (\%)},
        ylabel style={yshift=-3.em, scale=1.2},
        legend style={at={(0.5,1.2)}, anchor=north east, font={\tiny}, cells={anchor=west}, fill opacity=0.8, scale=1.2, legend columns=3}
        ]

        \addplot[purple,mark=square*,mark size=2.5pt,thick,mark options={fill=white,draw=purple,line width=1pt}] coordinates {(400,71.80) (800,72.09) (1200,71.83) (1600,71.97) (2000,71.86)};

        \addplot[orange,mark=diamond*,mark size=3pt,thick,mark options={fill=white,draw=orange,line width=1pt}] coordinates {(400,75.78) (800,75.78) (1200,75.33) (1600,75.85) (2000,75.92)};
    \end{axis}

    \begin{axis}
    [
        anchor=north west,
        at={(-2.5em,-5em)},
        ymajorgrids,
        xmajorgrids,
        grid style=dashed,
        width=.31\textwidth,
        height=.26\textwidth,
        yticklabel style={/pgf/number format/precision=0,/pgf/number format/fixed zerofill,scale=1.0},
        xmax=2100,
        xmin=300,
        ymin=84,
        ymax=93,
        xtick={400,800,1200,1600,2000},
        xticklabels={300,600,900,1200,1500},
        ytick={84,88,92},
        xlabel={\scriptsize{(d) Grounding.}},
        xlabel style={scale=1.2, yshift=0.2em, xshift=0.1em},
        ylabel=\footnotesize{\scriptsize hit@1 (\%)},
        ylabel style={yshift=-3.em, scale=1.2},
        legend style={at={(0.5,1.2)}, anchor=north east, font={\tiny}, cells={anchor=west}, fill opacity=0.8, scale=1.0, legend columns=3}
        ]

        \addplot[purple,mark=square*,mark size=2.5pt,thick,mark options={fill=white,draw=purple,line width=1pt}] coordinates {(400,85.60) (800,85.73) (1200,86.20) (1600,86.23) (2000,85.63)};

        \addplot[orange,mark=diamond*,mark size=3pt,thick,mark options={fill=white,draw=orange,line width=1pt}] coordinates {(400,91.53) (800,90.33) (1200,91.38) (1600,91.25) (2000,91.53)};
    \end{axis}  
    \end{tikzpicture}
    \caption{\textcolor{darkred}{Sensitivity analysis of the warm-up duration $T_{\text{warmup}}$.} We evaluate the impact of $T_{\text{warmup}}$ on four meta-task categories using both 2B and 7B models.}
    \label{fig:sensitivity_T}
\end{figure*}
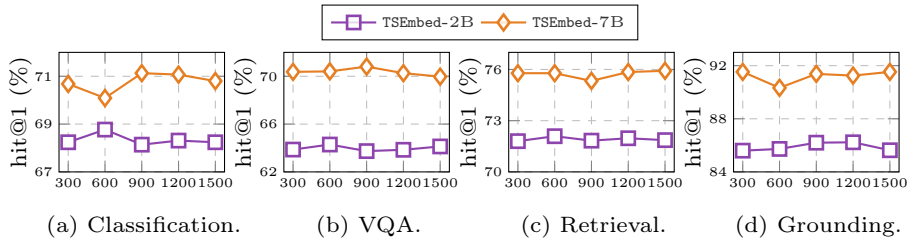

\vspace{1mm}
\noindent\textbf{Warm-Up Period ($T_{\text{warmup}}$).}
The parameter $T_{\text{warmup}}$ dictates when EANS activates, balancing router stabilization against refinement capacity. Figure~\ref{fig:sensitivity_T} reveals a capacity-dependent stabilization dynamic. The 2B model, with its smaller parameter space, rapidly establishes stable routing patterns, optimally converging at just 600 steps (peaking at 68.77\% for Classification, 64.29\% for VQA, 72.09\% for Retrieval). In contrast, the complex feature space of the 7B model demands a longer warm-up to solidify expert specialization, achieving optimal cross-task balance at 900--1200 steps, with Retrieval exhibiting a slight continued improvement up to 1,500 steps. Crucially, despite these distinct optimal points, both models exhibit relatively stable performance across adjacent step ranges, demonstrating the robustness of our two-stage learning paradigm.

\begin{table*}[!t]
	\centering
    \small
	\renewcommand{\arraystretch}{1.1}
    \caption{\textcolor{darkred}{Ablation study of \our.} \ding{51}/\ding{55} indicate enabled/disabled modules.}
    \vspace{-2mm}
	\label{tab:ablation_exp-mixtures-2}
    
    \resizebox{0.9\textwidth}{!}{
	\begin{tabular}{l ccc c cccc c}
		\toprule
		\multirow{2}{*}{\textbf{Configuration}} & \multicolumn{3}{c}{\textbf{Components}} & & \multicolumn{4}{c}{\textbf{Per Meta-Task Score}} & \multirow{2}{*}{\textbf{Overall}} \\
        \cmidrule(lr){2-4} \cmidrule(lr){6-9}
         & \textbf{MoE} & \textbf{EANS} & \textbf{Two-Stage} & & \textbf{Cls.} & \textbf{VQA}& \textbf{Ret.}& \textbf{Grd.} & \\
         \midrule[1pt]
        & \multicolumn{8}{c}{\textit{Qwen2-VL-2B~\cite{wang2024qwen2}}} \\
        \midrule[1pt]
		\rowcolor{gray!10}
        Baseline (VLM2VEC) & \ding{55} & \ding{55} & \ding{55} & & 59.00 & 49.40 & 65.40 & 73.40 & 59.30\\
        
        \rowcolor{white}
	    + MoE    & \ding{51} & \ding{55} & \ding{55} & & 67.87 & 64.17 & 72.03 & 85.65 & 70.20 \\

        \rowcolor{white}
		+ MoE + EANS & \ding{51} & \ding{51} & \ding{55} & & 68.31 & 64.26 & 71.59 & 85.08 & 70.14 \\
        
        \rowcolor{blue!5}
        \textbf{\our (Full)} & \ding{51} & \ding{51} & \ding{51} & & \textbf{68.77} & \textbf{64.29} & \textbf{72.09} & \textbf{85.73} & \textbf{70.52}\\

		\midrule[1pt]
        & \multicolumn{8}{c}{\textit{Qwen2-VL-7B~\cite{wang2024qwen2}}} \\
        \midrule[1pt]
		\rowcolor{gray!10}
        Baseline (VLM2VEC) & \ding{55} & \ding{55} & \ding{55} & & 62.60 & 57.80 & 69.90 & 81.70 & 65.80 \\
        
        \rowcolor{white}
	    + MoE    & \ding{51} & \ding{55} & \ding{55} & & 70.25 & 70.15 & 75.22 & 91.23 & 74.21 \\

        \rowcolor{white}
		+ MoE + EANS & \ding{51} & \ding{51} & \ding{55} & & 69.99 & 70.16 & 75.33 & 91.20 & 74.18 \\
        
        \rowcolor{blue!5}
        \textbf{\our (Full)} & \ding{51} & \ding{51} & \ding{51} & & \textbf{71.07} & \textbf{70.26} & \textbf{75.85} & \textbf{91.25} & \textbf{74.68}\\
		\bottomrule
	\end{tabular}
    }
\end{table*}

\vspace{-1mm}
\subsection{Ablation Study}

Table~\ref{tab:ablation_exp-mixtures-2} presents a detailed ablation analysis of \our. This breakdown analysis reveals three critical insights: \textbf{(1) MoE-LoRA effectively mitigates the problem of task conflict:} Transitioning from dense adaptation (VLM2VEC) to MoE-LoRA yields massive initial gains (+10.9\% at 2B, +8.41\% at 7B), proving that parameter-level expert specialization is essential to disentangle multimodal semantics. \textbf{(2) EANS strictly necessitates the two-stage warm-up:} Crucially, applying EANS without the two-stage paradigm marginally \emph{degrades} performance (e.g., dropping from 70.20\% to 70.14\% at 2B). This validates our hypothesis that premature, unstabilized routing distributions inject unreliable noise rather than helpful hard-negative signals. \textbf{(3) Component synergy unlocks peak performance:} Only when the two-stage paradigm first establishes a stable routing topology does EANS function as intended. This indispensable synergy yields consistent improvements across all meta-tasks, pushing the full \our framework to a peak performance of 70.52\% (2B) and 74.68\% (7B).

\subsection{More Analysis}

\definecolor{red}{RGB}{172,21,28}
\definecolor{blue}{RGB}{39,89,167}
\definecolor{red1}{RGB}{203,104,104}
\definecolor{blue1}{RGB}{104,155,203}

\definecolor{color4}{HTML}{015697}
\definecolor{color3}{HTML}{019496}
\definecolor{color2}{HTML}{F39C6B}
\definecolor{color1}{HTML}{FCAEA1}

\begin{figure*}[t]
\centering
\hspace{-2mm}
\begin{tikzpicture}
    \tiny{
  \begin{axis}[
    at={(-40em,-15.5em)},
    anchor=south west,
    ymajorgrids,
    grid style=dashed,
    legend style={at={(0.5,1)}, anchor=south west},
    legend cell align={left},
    ybar,
    enlarge x limits=0.65,
    xtick align=inside,
    width=.31\textwidth,
    height=.26\textwidth,
    bar width=0.65em,
    xlabel={\scriptsize{(a) Classification.}},
    xlabel style={scale=1.2, yshift=0.2em, xshift=-0.5em},
    ylabel=\footnotesize{\scriptsize hit@1 (\%)},
    ylabel style={yshift=-3.em, scale=1.2},
    symbolic x coords={{1}, {2},},
    xtick=data,
    ymin=60,
    ymax=72,
    ytick={60,66,72},
    nodes near coords align={vertical},
    xticklabels={2B, 7B},
    yticklabel style={/pgf/number format/fixed,/pgf/number format/fixed zerofill,/pgf/number format/precision=0,rotate=0,scale=1.0},
    legend style={yshift=0.8em,xshift=5.7em,font={\tiny},cells={anchor=west},fill opacity=0.8, scale=0.5, legend columns=4, font=\small}
    ]
    \addplot[fill=color1, draw=black, line width=0.6pt, area legend] coordinates {({1},68.18) ({2},70.94)};
    \addlegendentry{\scalebox{1.0}{{$N$=2}}}
    \addplot[fill=color2, draw=black, line width=0.6pt, area legend] coordinates {({1},68.77) ({2},71.07)};
    \addlegendentry{\scalebox{1.0}{{$N$=4}}}
    \addplot[fill=color3, draw=black, line width=0.6pt, area legend] coordinates {({1},68.51) ({2},71.06)};
    \addlegendentry{\scalebox{1.0}{{$N$=6}}}
    \addplot[fill=color4, draw=black, line width=0.6pt, area legend] coordinates {({1},67.99) ({2},70.81)};
    \addlegendentry{\scalebox{1.0}{{$N$=8}}}
  \end{axis}
	
  \begin{axis}[
    at={(-27.5em,-15.5em)},
    anchor=south west,
    ymajorgrids,
    grid style=dashed,
    legend style={at={(0.02,1)}, anchor=south west},
    legend cell align={left},
    ybar,
    enlarge x limits=0.65,
    xtick align=inside,
    width=.31\textwidth,
    height=.26\textwidth,
    bar width=0.65em,
    xlabel={\scriptsize{(b) VQA.}},
    xlabel style={scale=1.2, yshift=0.2em, xshift=-0.5em},
    ylabel=\footnotesize{\scriptsize hit@1 (\%)},
    ylabel style={yshift=-3.em, scale=1.2},
    symbolic x coords={{1}, {2},},
    xtick=data,
    ymin=60,
    ymax=72,
    ytick={60,66,72},
    nodes near coords align={vertical},
    xticklabels={2B, 7B},
    yticklabel style={/pgf/number format/fixed,/pgf/number format/fixed zerofill,/pgf/number format/precision=0,rotate=0,scale=1.0},
    legend style={yshift=0.2em,xshift=4.2em,font={\tiny},cells={anchor=west},fill opacity=0.8, scale=1.0, legend columns=3}
    ]
    \addplot[fill=color1, draw=black, line width=0.6pt, area legend] coordinates {({1},63.78) ({2},69.64)};
    \addplot[fill=color2, draw=black, line width=0.6pt, area legend] coordinates {({1},64.29) ({2},70.26)};
    \addplot[fill=color3, draw=black, line width=0.6pt, area legend] coordinates {({1},63.93) ({2},70.39)};
    \addplot[fill=color4, draw=black, line width=0.6pt, area legend] coordinates {({1},64.18) ({2},69.83)};
  \end{axis}

  \begin{axis}[
    at={(-15em,-15.5em)},
    anchor=south west,
    ymajorgrids,
    grid style=dashed,
    legend style={at={(0.02,1)}, anchor=south west},
    legend cell align={left},
    ybar,
    enlarge x limits=0.65,
    xtick align=inside,
    width=.31\textwidth,
    height=.26\textwidth,
    bar width=0.65em,
    xlabel={\scriptsize{(c) Retrieval.}},
    xlabel style={scale=1.2, yshift=0.2em, xshift=-0.5em},
    ylabel=\footnotesize{\scriptsize hit@1 (\%)},
    ylabel style={yshift=-3.em, scale=1.2},
    symbolic x coords={{1}, {2},},
    xtick=data,
    ymin=68,
    ymax=78,
    ytick={68,73,78},
    nodes near coords align={vertical},
    xticklabels={2B, 7B},
    yticklabel style={/pgf/number format/fixed,/pgf/number format/fixed zerofill,/pgf/number format/precision=0,rotate=0,scale=1.0},
    legend style={yshift=0.2em,xshift=4.2em,font={\tiny},cells={anchor=west},fill opacity=0.8, scale=1.0, legend columns=3}
    ]
    \addplot[fill=color1, draw=black, line width=0.6pt, area legend] coordinates {({1},72.14) ({2},75.50)};
    \addplot[fill=color2, draw=black, line width=0.6pt, area legend] coordinates {({1},72.09) ({2},75.85)};
    \addplot[fill=color3, draw=black, line width=0.6pt, area legend] coordinates {({1},72.14) ({2},75.97)};
    \addplot[fill=color4, draw=black, line width=0.6pt, area legend] coordinates {({1},71.78) ({2},75.50)};
  \end{axis}

  \begin{axis}[
    at={(-2.5em,-15.5em)},
    anchor=south west,
    ymajorgrids,
    grid style=dashed,
    legend style={at={(0.02,1)}, anchor=south west},
    legend cell align={left},
    ybar,
    enlarge x limits=0.65,
    xtick align=inside,
    width=.31\textwidth,
    height=.26\textwidth,
    bar width=0.65em,
    xlabel={\scriptsize{(d) Grounding.}},
    xlabel style={scale=1.2, yshift=0.2em, xshift=-0.5em},
    ylabel=\footnotesize{\scriptsize hit@1 (\%)},
    ylabel style={yshift=-3.em, scale=1.2},
    symbolic x coords={{1}, {2},},
    xtick=data,
    ymin=84,
    ymax=93,
    ytick={84,88,92},
    nodes near coords align={vertical},
    xticklabels={2B, 7B},
    yticklabel style={/pgf/number format/fixed,/pgf/number format/fixed zerofill,/pgf/number format/precision=0,rotate=0,scale=1.0},
    legend style={yshift=0.2em,xshift=4.2em,font={\tiny},cells={anchor=west},fill opacity=0.8, scale=1.0, legend columns=3}
    ]
    \addplot[fill=color1, draw=black, line width=0.6pt, area legend] coordinates {({1},85.40) ({2},90.98)};
    \addplot[fill=color2, draw=black, line width=0.6pt, area legend] coordinates {({1},85.73) ({2},91.25)};
    \addplot[fill=color3, draw=black, line width=0.6pt, area legend] coordinates {({1},85.98) ({2},91.20)};
    \addplot[fill=color4, draw=black, line width=0.6pt, area legend] coordinates {({1},86.03) ({2},90.75)};
  \end{axis}

}   
\end{tikzpicture}
\caption{\textcolor{darkred}{Impact of expert number $N$ on model performance.} We evaluate $N \in \{2, 4, 6, 8\}$ across four meta-task categories on both 2B and 7B models.}
\label{fig:expert_analysis}
\end{figure*}
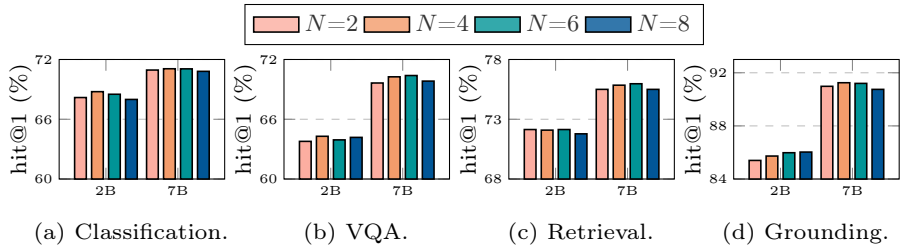
\definecolor{red}{RGB}{172,21,28}
\definecolor{blue}{RGB}{39,89,167}
\definecolor{red1}{RGB}{203,104,104}
\definecolor{blue1}{RGB}{104,155,203}

\definecolor{layer1}{RGB}{254,224,210}
\definecolor{layer2}{RGB}{252,146,114}
\definecolor{layer3}{RGB}{222,45,38}
\definecolor{layer4}{RGB}{165,15,21}

\definecolor{layer1}{HTML}{BFDFD2}
\definecolor{layer2}{HTML}{51999F}
\definecolor{layer3}{HTML}{ECB66C}
\definecolor{layer4}{HTML}{ED8D5A}

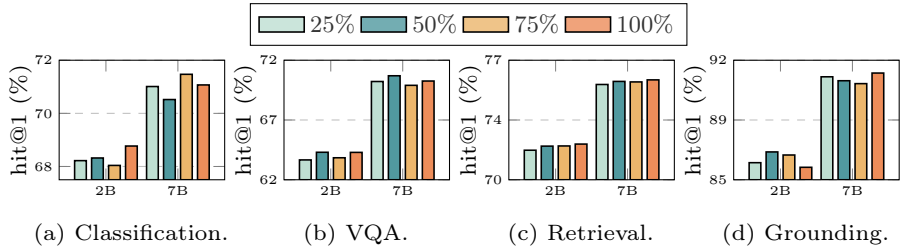
\begin{figure*}[t]
\centering
\hspace{-2mm}
\begin{tikzpicture}
    \tiny{
  \begin{axis}[
    at={(-40em,-15.5em)},
    anchor=south west,
    ymajorgrids,
    grid style=dashed,
    legend style={at={(0.5,1)}, anchor=south west},
    legend cell align={left},
    ybar,
    enlarge x limits=0.65,
    xtick align=inside,
    width=.31\textwidth,
    height=.26\textwidth,
    bar width=0.65em,
    xlabel={\scriptsize{(a) Classification.}},
    xlabel style={scale=1.2, yshift=0.2em, xshift=-0.5em},
    ylabel=\footnotesize{\scriptsize hit@1 (\%)},
    ylabel style={yshift=-3.em, scale=1.2},
    symbolic x coords={{1}, {2},},
    xtick=data,
    ymin=67.5,
    ymax=72,
    ytick={68,70,72},
    nodes near coords align={vertical},
    xticklabels={2B, 7B},
    yticklabel style={/pgf/number format/fixed,/pgf/number format/fixed zerofill,/pgf/number format/precision=0,rotate=0,scale=1.0},
    legend style={yshift=0.8em,xshift=6em,font={\tiny},cells={anchor=west},fill opacity=0.8, scale=0.5, legend columns=4, font=\small}
    ]
    \addplot[fill=layer1, draw=black, line width=0.6pt, area legend] coordinates {({1},68.22) ({2},71.01)};
    \addlegendentry{\scalebox{1.0}{{25\%}}}
    \addplot[fill=layer2, draw=black, line width=0.6pt, area legend] coordinates {({1},68.32) ({2},70.52)};
    \addlegendentry{\scalebox{1.0}{{50\%}}}
    \addplot[fill=layer3, draw=black, line width=0.6pt, area legend] coordinates {({1},68.04) ({2},71.47)};
    \addlegendentry{\scalebox{1.0}{{75\%}}}
    \addplot[fill=layer4, draw=black, line width=0.6pt, area legend] coordinates {({1},68.77) ({2},71.07)};
    \addlegendentry{\scalebox{1.0}{{100\%}}}
  \end{axis}
	
  \begin{axis}[
    at={(-27.5em,-15.5em)},
    anchor=south west,
    ymajorgrids,
    grid style=dashed,
    legend style={at={(0.02,1)}, anchor=south west},
    legend cell align={left},
    ybar,
    enlarge x limits=0.65,
    xtick align=inside,
    width=.31\textwidth,
    height=.26\textwidth,
    bar width=0.65em,
    xlabel={\scriptsize{(b) VQA.}},
    xlabel style={scale=1.2, yshift=0.2em, xshift=-0.5em},
    ylabel=\footnotesize{\scriptsize hit@1 (\%)},
    ylabel style={yshift=-3.em, scale=1.2},
    symbolic x coords={{1}, {2},},
    xtick=data,
    ymin=62,
    ymax=72,
    ytick={62,67,72},
    nodes near coords align={vertical},
    xticklabels={2B, 7B},
    yticklabel style={/pgf/number format/fixed,/pgf/number format/fixed zerofill,/pgf/number format/precision=0,rotate=0,scale=1.0},
    legend style={yshift=0.2em,xshift=4.2em,font={\tiny},cells={anchor=west},fill opacity=0.8, scale=1.0, legend columns=3}
    ]
    \addplot[fill=layer1, draw=black, line width=0.6pt, area legend] coordinates {({1},63.66) ({2},70.22)};
    \addplot[fill=layer2, draw=black, line width=0.6pt, area legend] coordinates {({1},64.30) ({2},70.70)};
    \addplot[fill=layer3, draw=black, line width=0.6pt, area legend] coordinates {({1},63.84) ({2},69.90)};
    \addplot[fill=layer4, draw=black, line width=0.6pt, area legend] coordinates {({1},64.29) ({2},70.26)};
  \end{axis}

  \begin{axis}[
    at={(-15em,-15.5em)},
    anchor=south west,
    ymajorgrids,
    grid style=dashed,
    legend style={at={(0.02,1)}, anchor=south west},
    legend cell align={left},
    ybar,
    enlarge x limits=0.65,
    xtick align=inside,
    width=.31\textwidth,
    height=.26\textwidth,
    bar width=0.65em,
    xlabel={\scriptsize{(c) Retrieval.}},
    xlabel style={scale=1.2, yshift=0.2em, xshift=-0.5em},
    ylabel=\footnotesize{\scriptsize hit@1 (\%)},
    ylabel style={yshift=-3.em, scale=1.2},
    symbolic x coords={{1}, {2},},
    xtick=data,
    ymin=70,
    ymax=77,
    ytick={70,73.5,77},
    nodes near coords align={vertical},
    xticklabels={2B, 7B},
    yticklabel style={/pgf/number format/fixed,/pgf/number format/fixed zerofill,/pgf/number format/precision=0,rotate=0,scale=1.0},
    legend style={yshift=0.2em,xshift=4.2em,font={\tiny},cells={anchor=west},fill opacity=0.8, scale=1.0, legend columns=3}
    ]
    \addplot[fill=layer1, draw=black, line width=0.6pt, area legend] coordinates {({1},71.73) ({2},75.58)};
    \addplot[fill=layer2, draw=black, line width=0.6pt, area legend] coordinates {({1},71.97) ({2},75.76)};
    \addplot[fill=layer3, draw=black, line width=0.6pt, area legend] coordinates {({1},71.98) ({2},75.73)};
    \addplot[fill=layer4, draw=black, line width=0.6pt, area legend] coordinates {({1},72.09) ({2},75.85)};
  \end{axis}

  \begin{axis}[
    at={(-2.5em,-15.5em)},
    anchor=south west,
    ymajorgrids,
    grid style=dashed,
    legend style={at={(0.02,1)}, anchor=south west},
    legend cell align={left},
    ybar,
    enlarge x limits=0.65,
    xtick align=inside,
    width=.31\textwidth,
    height=.26\textwidth,
    bar width=0.65em,
    xlabel={\scriptsize{(d) Grounding.}},
    xlabel style={scale=1.2, yshift=0.2em, xshift=-0.5em},
    ylabel=\footnotesize{\scriptsize hit@1 (\%)},
    ylabel style={yshift=-3.em, scale=1.2},
    symbolic x coords={{1}, {2},},
    xtick=data,
    ymin=85,
    ymax=92,
    ytick={85,88.5,92},
    nodes near coords align={vertical},
    xticklabels={2B, 7B},
    yticklabel style={/pgf/number format/fixed,/pgf/number format/fixed zerofill,/pgf/number format/precision=0,rotate=0,scale=1.0},
    legend style={yshift=0.2em,xshift=4.2em,font={\tiny},cells={anchor=west},fill opacity=0.8, scale=1.0, legend columns=3}
    ]
    \addplot[fill=layer1, draw=black, line width=0.6pt, area legend] coordinates {({1},86.00) ({2},91.03)};
    \addplot[fill=layer2, draw=black, line width=0.6pt, area legend] coordinates {({1},86.63) ({2},90.80)};
    \addplot[fill=layer3, draw=black, line width=0.6pt, area legend] coordinates {({1},86.45) ({2},90.63)};
    \addplot[fill=layer4, draw=black, line width=0.6pt, area legend] coordinates {({1},85.73) ({2},91.25)};
  \end{axis}

}   
\end{tikzpicture}
\caption{\textcolor{darkred}{Impact of layer coverage on EANS routing aggregation.} We evaluate the effect of aggregating router distributions from 25\%, 50\%, 75\%, and 100\% of layers.}
\label{fig:EANS_layers}
\end{figure*}

\noindent\textbf{Impact of Expert Number.}
As illustrated in Figure~\ref{fig:expert_analysis}, we sweep the number of experts $N \in \{2, 4, 6, 8\}$ to evaluate scaling behaviors. The results reveal a clear performance "sweet spot" at $N=4$, which achieves peak or near-peak performance. We attribute this to an organic alignment between the network capacity and the underlying taxonomy of the MMEB benchmark, which inherently comprises four distinct meta-task clusters. A smaller $N$ (e.g., $N=2$) forces disparate tasks into shared parameters, failing to fully disentangle conflicts. Conversely, scaling to $N=8$ generally degrades performance, suggesting that an excessive number of experts over-fragments the semantic space, introducing severe routing ambiguity that outweighs any marginal capacity gains. While $N=6$ offers slight domain-specific advantages (e.g., 75.97\% on 7B retrieval), $N=4$ yields the most robust cross-task generalization. Based on these findings, we recommend configuring $N$ to roughly match the number of macro-task clusters in the target domain to prevent semantic over-partitioning.

\vspace{1mm}
\noindent\textbf{Impact of Layer Coverage for EANS.}
Figure~\ref{fig:EANS_layers} demonstrates the effect of varying the EANS routing aggregation coverage from 25\% to 100\% of the backbone layers. 
The benefit of full coverage is explicitly evident in retrieval, which exhibits a near-monotonic performance trajectory, culminating at 72.09\% (2B) and 75.85\% (7B).
This confirms that holistic, cross-layer routing effectively amalgamates the low-level and high-level semantic signals necessary for discerning hard negatives. While intermediate coverage levels (e.g., 50\%) occasionally yield slight localized peaks in VQA and grounding, 100\% coverage consistently maintains robust, top-tier performance across all meta-tasks and model scales.

\section{Related Work}
\label{sec:related}


\noindent\textbf{Universal Multimodal Embedding.} Early dual-encoder models, such as CLIP~\citep{radford2021learning} and SigLIP~\citep{tschannen2025siglip}, excel at coarse-grained alignment but struggle with complex reasoning due to their shallow cross-modal interactions. Recently, the proliferation of MLLMs has catalyzed a paradigm shift toward unified generative embeddings. For instance, Qwen3-VL-Embedding~\citep{li2026qwen3} leverages its foundational model to synthesize task-specific multimodal annotations from a curated seed pool, augmented by positive refinement and hard-negative mining. Concurrently, QQMM-embed~\citep{xue2025improve} explicitly amplifies the gradients of hard negatives to enforce sharper discriminative boundaries. From the reasoning perspective, TTE~\citep{cui2025think} employs a MLLM to generate logical traces, conditioning the final embeddings on these intermediate explanatory steps. 
However, these methods suffer from severe task conflict, which \our is explicitly designed to address.


\vspace{1.2mm}
\noindent\textbf{Task Conflict in Multi-Task Learning.}
Task conflict, where gradients from divergent objectives destructively interfere, is a fundamental bottleneck in multi-task learning~\citep{zhang2021survey,chen2024multi}. Traditional mitigations typically involve explicit gradient manipulation (e.g., PCGrad~\citep{yu2020gradient}, GradNorm~\citep{chen2018gradnorm}, Nash-MTL~\citep{navon2022multi}) or static architecture routing (e.g., Cross-stitch~\citep{misra2016cross}, AdaShare~\citep{sun2020adashare}). While effective for small, fixed task sets, these methods become computationally prohibitive and structurally intractable when scaled to the massive, open-ended distributions of universal multimodal embeddings. To address this, our MoE-based architecture provides an elegant alternative, achieving automatic task decomposition and gradient isolation without the exorbitant costs of explicit gradient engineering.

\vspace{1.2mm}
\noindent\textbf{Hard Negative Sampling.}
Negative sample selection critically dictates the efficacy of contrastive representation learning~\citep{chen2020simple,he2020momentum}. Standard approaches to hard negative mining often rely on maintaining large memory banks (e.g., MoCHi~\citep{kalantidis2020hard}), density-aware resampling~\citep{robinson2020contrastive}, or complex curriculum schedules~\citep{zhang2022dual}. While these methods effectively improve cluster separation, they inevitably introduce significant computational overhead by requiring explicit feature-level similarity computations or external memory mechanisms. In contrast, our EANS strategically repurposes the MoE routing mechanism as a free intrinsic semantic proxy. 
\section{Conclusion}
\label{sec:conclusion}

In this paper, we introduce \our to address the challenge of task conflict in universal multimodal embeddings. Our framework uniquely combines MoE with LoRA to achieve task-specific parameter specialization, effectively disentangling conflicting semantic objectives. Furthermore, we propose Expert-Aware Negative Sampling (EANS), a novel strategy that leverages routing distributions as an intrinsic proxy to dynamically emphasize hard negatives. To ensure training stability, we devise a two-stage learning paradigm that solidifies expert specialization before deploying EANS, guaranteeing robust optimization. Extensive evaluations on both the MMEB and real-world production datasets demonstrate the effectiveness of \our while preserving training efficiency.



%
%
\bibliographystyle{splncs04}
\bibliography{main}

\begin{thebibliography}{10}
\providecommand{\url}[1]{\texttt{#1}}
\providecommand{\urlprefix}{URL }
\providecommand{\doi}[1]{https://doi.org/#1}

\bibitem{cao2022image}
Cao, M., Li, S., Li, J., Nie, L., Zhang, M.: Image-text retrieval: A survey on recent research and development. arXiv preprint arXiv:2203.14713  (2022)

\bibitem{chen2024multi}
Chen, S., Zhang, Y., Yang, Q.: Multi-task learning in natural language processing: An overview. ACM Computing Surveys  \textbf{56}(12),  1--32 (2024)

\bibitem{chen2020simple}
Chen, T., Kornblith, S., Norouzi, M., Hinton, G.: A simple framework for contrastive learning of visual representations. In: International conference on machine learning. pp. 1597--1607. PmLR (2020)

\bibitem{chen2018gradnorm}
Chen, Z., Badrinarayanan, V., Lee, C.Y., Rabinovich, A.: Gradnorm: Gradient normalization for adaptive loss balancing in deep multitask networks. In: International conference on machine learning. pp. 794--803. PMLR (2018)

\bibitem{cherti2023reproducible}
Cherti, M., Beaumont, R., Wightman, R., Wortsman, M., Ilharco, G., Gordon, C., Schuhmann, C., Schmidt, L., Jitsev, J.: Reproducible scaling laws for contrastive language-image learning. In: Proceedings of the IEEE/CVF conference on computer vision and pattern recognition. pp. 2818--2829 (2023)

\bibitem{cui2025think}
Cui, X., Cheng, J., Chen, H.y., Shukla, S.N., Awasthi, A., Pan, X., Ahuja, C., Mishra, S.K., Yang, Y., Xiao, J., et~al.: Think then embed: Generative context improves multimodal embedding. arXiv preprint arXiv:2510.05014  (2025)

\bibitem{faysse2024colpali}
Faysse, M., Sibille, H., Wu, T., Omrani, B., Viaud, G., Hudelot, C., Colombo, P.: Colpali: Efficient document retrieval with vision language models. arXiv preprint arXiv:2407.01449  (2024)

\bibitem{gao2021scaling}
Gao, L., Zhang, Y., Han, J., Callan, J.: Scaling deep contrastive learning batch size under memory limited setup. arXiv preprint arXiv:2101.06983  (2021)

\bibitem{greenacre2022principal}
Greenacre, M., Groenen, P.J., Hastie, T., d’Enza, A.I., Markos, A., Tuzhilina, E.: Principal component analysis. Nature Reviews Methods Primers  \textbf{2}(1), ~100 (2022)

\bibitem{gu2025unime}
Gu, T., Yang, K., Zhang, K., An, X., Feng, Z., Zhang, Y., Cai, W., Deng, J., Bing, L.: Unime-v2: Mllm-as-a-judge for universal multimodal embedding learning. arXiv preprint arXiv:2510.13515  (2025)

\bibitem{he2020momentum}
He, K., Fan, H., Wu, Y., Xie, S., Girshick, R.: Momentum contrast for unsupervised visual representation learning. In: Proceedings of the IEEE/CVF conference on computer vision and pattern recognition. pp. 9729--9738 (2020)

\bibitem{hu2022lora}
Hu, E.J., Shen, Y., Wallis, P., Allen-Zhu, Z., Li, Y., Wang, S., Wang, L., Chen, W., et~al.: Lora: Low-rank adaptation of large language models. ICLR  \textbf{1}(2), ~3 (2022)

\bibitem{jian2025rzenembed}
Jian, W., Zhang, Y., Liang, D., Xie, C., He, Y., Leng, D., Yin, Y.: Rzenembed: Towards comprehensive multimodal retrieval. arXiv preprint arXiv:2510.27350  (2025)

\bibitem{jiang2026embed}
Jiang, H., Wang, Y., Zhu, Y., Lu, X., Qin, W., Wang, M., Wan, P., Tang, Y.: Embed-rl: Reinforcement learning for reasoning-driven multimodal embeddings. arXiv preprint arXiv:2602.13823  (2026)

\bibitem{jiang2024e5}
Jiang, T., Song, M., Zhang, Z., Huang, H., Deng, W., Sun, F., Zhang, Q., Wang, D., Zhuang, F.: E5-v: Universal embeddings with multimodal large language models. arXiv preprint arXiv:2407.12580  (2024)

\bibitem{jiang2024vlm2vec}
Jiang, Z., Meng, R., Yang, X., Yavuz, S., Zhou, Y., Chen, W.: Vlm2vec: Training vision-language models for massive multimodal embedding tasks. arXiv preprint arXiv:2410.05160  (2024)

\bibitem{kalantidis2020hard}
Kalantidis, Y., Sariyildiz, M.B., Pion, N., Weinzaepfel, P., Larlus, D.: Hard negative mixing for contrastive learning. Advances in neural information processing systems  \textbf{33},  21798--21809 (2020)

\bibitem{kong2025modality}
Kong, F., Zhang, J., Liu, Y., Zhang, H., Feng, S., Yang, X., Wang, D., Tian, Y., Zhang, F., Zhou, G., et~al.: Modality curation: Building universal embeddings for advanced multimodal information retrieval. arXiv preprint arXiv:2505.19650  (2025)

\bibitem{lan2025llave}
Lan, Z., Niu, L., Meng, F., Zhou, J., Su, J.: Llave: Large language and vision embedding models with hardness-weighted contrastive learning. arXiv preprint arXiv:2503.04812  (2025)

\bibitem{li2023blip}
Li, J., Li, D., Savarese, S., Hoi, S.: Blip-2: Bootstrapping language-image pre-training with frozen image encoders and large language models. In: International conference on machine learning. pp. 19730--19742. PMLR (2023)

\bibitem{li2026qwen3}
Li, M., Zhang, Y., Long, D., Chen, K., Song, S., Bai, S., Yang, Z., Xie, P., Yang, A., Liu, D., et~al.: Qwen3-vl-embedding and qwen3-vl-reranker: A unified framework for state-of-the-art multimodal retrieval and ranking. arXiv preprint arXiv:2601.04720  (2026)

\bibitem{liu2024multimodal}
Liu, Q., Zhu, J., Yang, Y., Dai, Q., Du, Z., Wu, X.M., Zhao, Z., Zhang, R., Dong, Z.: Multimodal pretraining, adaptation, and generation for recommendation: A survey. In: Proceedings of the 30th ACM SIGKDD Conference on Knowledge Discovery and Data Mining. pp. 6566--6576 (2024)

\bibitem{liu2025lamra}
Liu, Y., Zhang, Y., Cai, J., Jiang, X., Hu, Y., Yao, J., Wang, Y., Xie, W.: Lamra: Large multimodal model as your advanced retrieval assistant. In: Proceedings of the Computer Vision and Pattern Recognition Conference. pp. 4015--4025 (2025)

\bibitem{meng2025vlm2vec}
Meng, R., Jiang, Z., Liu, Y., Su, M., Yang, X., Fu, Y., Qin, C., Chen, Z., Xu, R., Xiong, C., et~al.: Vlm2vec-v2: Advancing multimodal embedding for videos, images, and visual documents. arXiv preprint arXiv:2507.04590  (2025)

\bibitem{misra2016cross}
Misra, I., Shrivastava, A., Gupta, A., Hebert, M.: Cross-stitch networks for multi-task learning. In: Proceedings of the IEEE conference on computer vision and pattern recognition. pp. 3994--4003 (2016)

\bibitem{navon2022multi}
Navon, A., Shamsian, A., Achituve, I., Maron, H., Kawaguchi, K., Chechik, G., Fetaya, E.: Multi-task learning as a bargaining game. arXiv preprint arXiv:2202.01017  (2022)

\bibitem{oord2018representation}
Oord, A.v.d., Li, Y., Vinyals, O.: Representation learning with contrastive predictive coding. arXiv preprint arXiv:1807.03748  (2018)

\bibitem{radford2021learning}
Radford, A., Kim, J.W., Hallacy, C., Ramesh, A., Goh, G., Agarwal, S., Sastry, G., Askell, A., Mishkin, P., Clark, J., et~al.: Learning transferable visual models from natural language supervision. In: International conference on machine learning. pp. 8748--8763. PmLR (2021)

\bibitem{robinson2020contrastive}
Robinson, J., Chuang, C.Y., Sra, S., Jegelka, S.: Contrastive learning with hard negative samples. arXiv preprint arXiv:2010.04592  (2020)

\bibitem{song2025comprehensive}
Song, X., Lin, H., Wen, H., Hou, B., Xu, M., Nie, L.: A comprehensive survey on composed image retrieval. ACM Transactions on Information Systems  \textbf{44}(1),  1--54 (2025)

\bibitem{sun2020adashare}
Sun, X., Panda, R., Feris, R., Saenko, K.: Adashare: Learning what to share for efficient deep multi-task learning. Advances in Neural Information Processing Systems  \textbf{33},  8728--8740 (2020)

\bibitem{thirukovalluru2025breaking}
Thirukovalluru, R., Meng, R., Liu, Y., Su, M., Nie, P., Yavuz, S., Zhou, Y., Chen, W., Dhingra, B., et~al.: Breaking the batch barrier (b3) of contrastive learning via smart batch mining. arXiv preprint arXiv:2505.11293  (2025)

\bibitem{tschannen2025siglip}
Tschannen, M., Gritsenko, A., Wang, X., Naeem, M.F., Alabdulmohsin, I., Parthasarathy, N., Evans, T., Beyer, L., Xia, Y., Mustafa, B., et~al.: Siglip 2: Multilingual vision-language encoders with improved semantic understanding, localization, and dense features. arXiv preprint arXiv:2502.14786  (2025)

\bibitem{wang2024qwen2vlenhancingvisionlanguagemodels}
Wang, P., Bai, S., Tan, S., Wang, S., Fan, Z., Bai, J., Chen, K., Liu, X., Wang, J., Ge, W., Fan, Y., Dang, K., Du, M., Ren, X., Men, R., Liu, D., Zhou, C., Zhou, J., Lin, J.: Qwen2-vl: Enhancing vision-language model's perception of the world at any resolution (2024), \url{https://arxiv.org/abs/2409.12191}

\bibitem{wang2024qwen2}
Wang, P., Bai, S., Tan, S., Wang, S., Fan, Z., Bai, J., Chen, K., Liu, X., Wang, J., Ge, W., et~al.: Qwen2-vl: Enhancing vision-language model's perception of the world at any resolution. arXiv preprint arXiv:2409.12191  (2024)

\bibitem{wei2024uniir}
Wei, C., Chen, Y., Chen, H., Hu, H., Zhang, G., Fu, J., Ritter, A., Chen, W.: Uniir: Training and benchmarking universal multimodal information retrievers. In: European Conference on Computer Vision. pp. 387--404. Springer (2024)

\bibitem{wu2025elastic}
Wu, Y., Li, J., Guo, Z., Li, L.: Elastic mixture of rank-wise experts for knowledge reuse in federated fine-tuning. arXiv preprint arXiv:2512.00902  (2025)

\bibitem{wu2025learning}
Wu, Y., Li, J., Guo, Z., Li, L.: Learning like humans: Resource-efficient federated fine-tuning through cognitive developmental stages. arXiv preprint arXiv:2508.00041  (2025)

\bibitem{wu2025memory}
Wu, Y., Li, J., Tian, C., Guo, Z., Li, L.: Memory-efficient federated fine-tuning of large language models via layer pruning. arXiv preprint arXiv:2508.17209  (2025)

\bibitem{xu2024copyrightmeter}
Xu, N., Li, C., Du, T., Li, M., Luo, W., Liang, J., Li, Y., Zhang, X., Han, M., Yin, J., et~al.: Copyrightmeter: Revisiting copyright protection in text-to-image models. arXiv preprint arXiv:2411.13144  (2024)

\bibitem{xu2025bridging}
Xu, N., Zhang, J., Li, C., An, H., Zhou, C., Wang, J., Xu, B., Li, Y., Du, T., Ji, S.: Bridging the copyright gap: Do large vision-language models recognize and respect copyrighted content? arXiv preprint arXiv:2512.21871  (2025)

\bibitem{xu2025videoeraser}
Xu, N., Zhang, J., Li, C., Chen, Z., Zhou, C., Li, Q., Du, T., Ji, S.: Videoeraser: Concept erasure in text-to-video diffusion models. In: Proceedings of the 2025 Conference on Empirical Methods in Natural Language Processing. pp. 5965--5994 (2025)

\bibitem{xue2025improve}
Xue, Y., Li, D., Liu, G.: Improve multi-modal embedding learning via explicit hard negative gradient amplifying. arXiv preprint arXiv:2506.02020  (2025)

\bibitem{yang2023dawn}
Yang, Z., Li, L., Lin, K., Wang, J., Lin, C.C., Liu, Z., Wang, L.: The dawn of lmms: Preliminary explorations with gpt-4v (ision). arXiv preprint arXiv:2309.17421  (2023)

\bibitem{ye2025harnessing}
Ye, Y., Zheng, Z., Shen, Y., Wang, T., Zhang, H., Zhu, P., Yu, R., Zhang, K., Xiong, H.: Harnessing multimodal large language models for multimodal sequential recommendation. In: Proceedings of the AAAI Conference on Artificial Intelligence. vol.~39, pp. 13069--13077 (2025)

\bibitem{yu2024evaluation}
Yu, H., Gan, A., Zhang, K., Tong, S., Liu, Q., Liu, Z.: Evaluation of retrieval-augmented generation: A survey. In: CCF Conference on Big Data. pp. 102--120. Springer (2024)

\bibitem{yu2025cafe}
Yu, H., Zhao, Z., Yan, S., Korycki, L., Wang, J., He, B., Liu, J., Zhang, L., Fan, X., Yu, H.: Cafe: Unifying representation and generation with contrastive-autoregressive finetuning. arXiv preprint arXiv:2503.19900  (2025)

\bibitem{yu2020gradient}
Yu, T., Kumar, S., Gupta, A., Levine, S., Hausman, K., Finn, C.: Gradient surgery for multi-task learning. Advances in neural information processing systems  \textbf{33},  5824--5836 (2020)

\bibitem{zhang2025notellm}
Zhang, C., Zhang, H., Wu, S., Wu, D., Xu, T., Zhao, X., Gao, Y., Hu, Y., Chen, E.: Notellm-2: Multimodal large representation models for recommendation. In: Proceedings of the 31st ACM SIGKDD Conference on Knowledge Discovery and Data Mining V. 1. pp. 2815--2826 (2025)

\bibitem{zhang2022dual}
Zhang, C., Zhang, K., Pham, T.X., Niu, A., Qiao, Z., Yoo, C.D., Kweon, I.S.: Dual temperature helps contrastive learning without many negative samples: Towards understanding and simplifying moco. In: Proceedings of the IEEE/CVF conference on computer vision and pattern recognition. pp. 14441--14450 (2022)

\bibitem{zhang2024magiclens}
Zhang, K., Luan, Y., Hu, H., Lee, K., Qiao, S., Chen, W., Su, Y., Chang, M.W.: Magiclens: Self-supervised image retrieval with open-ended instructions. arXiv preprint arXiv:2403.19651  (2024)

\bibitem{zhang2025bridging}
Zhang, X., Zhang, Y., Xie, W., Li, M., Dai, Z., Long, D., Xie, P., Zhang, M., Li, W., Zhang, M.: Bridging modalities: Improving universal multimodal retrieval by multimodal large language models. In: Proceedings of the Computer Vision and Pattern Recognition Conference. pp. 9274--9285 (2025)

\bibitem{zhang2021survey}
Zhang, Y., Yang, Q.: A survey on multi-task learning. IEEE transactions on knowledge and data engineering  \textbf{34}(12),  5586--5609 (2021)

\bibitem{zhao2026retrieval}
Zhao, P., Zhang, H., Yu, Q., Wang, Z., Geng, Y., Fu, F., Yang, L., Zhang, W., Jiang, J., Cui, B.: Retrieval-augmented generation for ai-generated content: A survey. Data Science and Engineering pp. 1--29 (2026)

\end{thebibliography}
\end{document}